\documentclass{clv3}

\usepackage{hyperref}
\usepackage{xcolor}
\definecolor{darkblue}{rgb}{0, 0, 0.5}
\hypersetup{colorlinks=true,citecolor=darkblue, linkcolor=darkblue, urlcolor=darkblue}

\usepackage{textcomp}
\usepackage[official]{eurosym}
\usepackage[british]{babel}
\usepackage{booktabs}
\usepackage{linguex}
\usepackage[normalem]{ulem}
\usepackage{todonotes} 
\usepackage{tikz-dependency}
\usepackage[utf8]{inputenc}

\issue{1}{1}{2019}

\dochead{}
\runningtitle{Robust Extraction of Potentially Idiomatic Expressions}
\runningauthor{Haagsma, Nissim \& Bos}
\begin{document}

\title{\textit{Casting a Wide Net}: Robust Extraction of Potentially Idiomatic Expressions}

\author{Hessel Haagsma\thanks{E-mail: \texttt{\{hessel.haagsma, m.nissim, johan.bos\}@rug.nl}}}
\affil{CLCG, University of Groningen}

\author{Malvina Nissim*}
\affil{CLCG, University of Groningen}

\author{Johan Bos*}
\affil{CLCG, University of Groningen}

\maketitle

\begin{abstract}
Idiomatic expressions like `out of the woods' and `up the ante' present a range of difficulties for natural language processing applications. We present work on the annotation and extraction of what we term potentially idiomatic expressions (PIEs), a subclass of multiword expressions covering both literal and non-literal uses of idiomatic expressions. Existing corpora of PIEs are small and have limited coverage of different PIE types, which hampers research. To further progress on the extraction and disambiguation of potentially idiomatic expressions, larger corpora of PIEs are required. In addition, larger corpora are a potential source for valuable linguistic insights into idiomatic expressions and their variability. We propose automatic tools to facilitate the building of larger PIE corpora, by investigating the feasibility of using dictio\-na\-ry-based extraction of PIEs as a pre-extraction tool for English. We do this by assessing the reliability and coverage of idiom dictionaries, the annotation of a PIE corpus, and the automatic extraction of PIEs from a large corpus. Results show that combinations of dictionaries are a reliable source of idiomatic expressions, that PIEs can be annotated with a high reliability (0.74-0.91 Fleiss' Kappa), and that parse-based PIE extraction yields highly accurate performance (88\% F1-score). Combining complementary PIE extraction methods increases reliability further, to over 92\% F1-score. Moreover, the extraction method presented here could be extended to other types of multiword expressions and to other languages, given that sufficient NLP tools are available.
\end{abstract}

\newpage
\section{Introduction}
Idiomatic expressions pose a major challenge for a wide range of applications in natural language processing \citep{Sag2002}. These include machine translation \citep{Salton2014,Isabelle2017}, semantic parsing \citep{Fischer1996}, sentiment analysis \citep{Williams2015}, and word sense disambiguation \citep{Finlayson2011}. Idioms show significant syntactic and morphological variability (e.g. \textit{beans being spilled} for \textit{spill the beans}), which makes them hard to find automatically. Moreover, their non-compositional nature makes idioms really hard to interpret, because their meaning is often very different from the meanings of the words that make them up. Hence, successful systems need not only be able to recognise idiomatic expressions in text or dialogue, but they also need to give a proper interpretation to them. As a matter of fact, current language technology performs badly on idiom understanding, a phenomenon that perhaps has not received enough attention.

Nearly all current language technology used in NLP applications is based on supervised machine learning. This requires large amounts of labelled data. In the case of idiom interpretation, however, only small datasets are available. These contain just a couple of thousand idiom instances, covering only about fifty different types of idiomatic expressions. In fact, existing annotated corpora tend to cover only a small set of idiom types, comprising just a few syntactic patterns (e.g., verb-object combinations), of which a limited number of instances are extracted from a large corpus. 

This is not surprising as preparing and compiling such corpora involves a large amount of manual extraction work, especially if one wants to allow for form variation in the idiomatic expressions (for example, extracting \textit{cooking all the books} for \textit{cook the books}). This work involves both the crafting of syntactic patterns to match potential idiomatic expressions and the filtering of false extractions (non-instances of the target expression e.g. due to wrong parses), and increases with the amount of idiom types included in the corpus (which, in the worst case, means an exponential increase in false extractions). Thus, building a large corpus of idioms, especially one that covers many types in many syntactic constructions, is costly. If a high-precision, high-recall system can be developed for the task of extracting the annotation candidates, this cost will be greatly reduced, making the construction of a large corpus much more feasible.

The variability of idioms has been a significant topic of interest among researchers of idioms. For example, \citet{Minugh2007} investigates the internal and external modification of a set of idioms in a large English corpus, whereas \citet{Gregoire2009}, quantifies and classifies the variation of a set of idioms in a large corpus of Dutch, setting up a useful taxonomy of variation types. Both find that, although idiomatic expressions mainly occur in their dictionary form, there is a significant minority of idiom instances that occur in non-dictionary variants. Additionally, \citet{Geeraert2017} show that idiom variants retain their idiomatic meaning more often and are processed more easily than previously assumed. This emphasises the need for corpora covering idiomatic expressions to include these variants, and for tools to be robust in dealing with them.

As such, the aim of this article is to describe methods and provide tools for constructing larger corpora annotated with a wider range of idiom types than currently in existence due to the reduced amount of manual labour required. In this way we hope to stimulate further research in this area. In contrast to previous approaches, we want to catch as many idiomatic expressions as possible, and we achieve this by \textit{casting a wide net}, that is, we consider the widest range of possible idiom variants first and then filter out any bycatch 
in a way that requires the least manual effort. 

We expect research will benefit from having larger corpora by improving evaluation quality, by allowing for the training of better supervised systems, and by providing additional linguistic insight into idiomatic expressions. A reliable method for extracting idiomatic expressions is not only needed for building an annotated corpus, but can also be used as part of an automatic idiom processing pipeline. In such a pipeline, extracting potentially idiomatic expressions can be seen as a first step before idiom disambiguation, and the combination of the two modules then functions as an complete idiom extraction system.

The main research question that we aim to answer in this article is whether dictionary-based extraction of potentially idiomatic expressions is robust and reliable enough to facilitate the creation of wide-coverage sense-annotated idiom corpora.


By answering this question we make several
contributions to research on multiword expressions, in particular that of idiom extraction. 
Firstly, we provide an overview of existing research on annotating idiomatic expressions in corpora, showing that current corpora cover only small sets of idiomatic types (Section~\ref{sec:prev-work}).
Secondly, we quantify the coverage and reliability of a set of idiom dictionaries, demonstrating that there is little overlap between resources (Section~\ref{sec:dictionary-comparison}).
Thirdly, we develop and release an evaluation corpus for extracting potentially idiomatic expressions from text  (Section~\ref{sec:corpus-annotation})\footnote{The corpus annotations are available at https://github.com/hslh/pie-annotation under a CC-BY-4.0 licence.}. 
Finally, various extraction systems and combinations thereof are implemented, made available to the research community, and evaluated empirically (Section~\ref{sec:pie-extraction}).\footnote{The source code of the PIE extraction system is available at https://github.com/hslh/pie-detection under a CC-BY-4.0 licence.}

\section{New Terminology: Potentially Idiomatic Expression (PIE)}
\label{sec:terminology}
The ambiguity of phrases like \textit{wake up and smell the coffee} poses a terminological problem. Usually, these phrases are called \textit{idiomatic expressions}, which is suitable when they are used in an idiomatic sense, but not so much when they are used in a literal sense. Therefore, we propose a new%
\footnote{Note that \citet{Cook2008} came up with a similar term, \textit{potentially-idiomatic combinations}.}
term: \textit{potentially idiomatic expressions}, or \textit{PIEs} for short. The term \textit{potentially idiomatic expression} refers to those expressions which can have an idiomatic meaning, regardless of whether they actually have that meaning in a given context.%
\footnote{Ambiguity is not equally distributed across phrases. As with words, there are single sense phrases, with only an idiomatic sense, such as \textit{piping hot}, which can only get the figurative interpretation `very hot'. More commonly, there are phrases with exactly two senses, a literal and an idiomatic sense, such as \textit{wake up and smell the coffee}, which can take the literal meaning of `waking up and smelling coffee', and the idiomatic meaning of `facing reality and stop deluding oneself'. Sometimes, phrases can have more than two senses, e.g. one literal sense and multiple idiomatic ones, as in \textit{fall by the wayside}, which can take the literal meaning `fall down by the side of the road', the idiomatic meaning `fail to persist in an endeavour', and the alternative idiomatic meaning `be left without help'.}
So, \textit{see the light} is a PIE in both `After another explanation, I finally saw the light' and `I saw the light of the sun through the trees', while it is an idiomatic expression in the first context, and a literal phrase in the latter context.

The processing of PIEs involves three main challenges: the discovery of (new) PIE types, the extraction of instances of known PIE types in text, and the disambiguation of PIE instances in context. Here, we propose calling the discovery task simply \textit{PIE discovery}, the extraction task simply \textit{PIE extraction}, and the disambiguation task \textit{PIE disambiguation}. Note that these terms contrast with the terms used in existing research. There, the discovery task is called \textit{type-based idiom detection} and the disambiguation task is called \textit{token-based idiom detection} (cf. \citealp[for example]{Sporleder2010,Gharbieh2016}), although this usage is not always consistent. Because these terms are very similar, they are potentially confusing, and that is why we propose novel terminology.

Other terminology comes from literature on multiword expressions (MWEs) more generally, i.e. not specific to idioms. Here, the task of finding new MWE types is called \textit{MWE discovery} and finding instances of known MWE types is called \textit{MWE identification} \cite{Constant2017}. Note, however, that \textit{MWE identification} generally consists of finding only the idiomatic usages of these
types (e.g. \citealp{Ramisch2018}). This means that \textit{MWE identification} consists of both the extraction and disambiguation tasks, performed jointly. In this work, we propose to split this into two separate tasks, and we are concerned only with the \textit{PIE extraction} part, leaving \textit{PIE disambiguation} as a separate problem.

\section{Related Work}
\label{sec:prev-work}
This section is structured so as to reflect the dual contribution of the present work. First, we discuss existing resources annotated for idiomatic expressions. Second, we discuss existing approaches to the automatic extraction of idioms.

\subsection{Annotated Corpora and Annotation Schemes for Idioms}
\label{sec:prev-work-corpora}
There are four sizeable sense-annotated PIE corpora for English: the VNC-Tokens Dataset \citep{Cook2008}, the Gigaword dataset \citep{Sporleder2009}, the IDIX Corpus \citep{Sporleder2010}, and the SemEval-2013 Task 5 dataset \citep{Korkontzelos2013}. An overview of these corpora is presented in Table~\ref{table:corpora}.

\begin{table}[hbtp]
	\centering
    \setlength{\tabcolsep}{3.5pt}
	\caption{Overview of existing corpora of potentially idiomatic expressions and sense annotations for English. `Min' indicates the count for the least frequent idiom type, `Med' the median, and `Max' the most frequent type. The syntax types column indicates the syntactic patterns of the idiom types included in the dataset. The base corpora are the British National Corpus (BNC, \citealp{BNCRef}), ukWaC \cite{UKWAC}, and Gigaword \cite{Gigaword2003}.}
	\label{table:corpora}
	\begin{tabular}{lrrrrrrll}
		\toprule
		Name & Types & Instances & Min & Med & Max & Senses & Corpus & Syntax Types\\
		\midrule
		VNC-Tokens & 53 & 2,984 & 21 & 49 & 108 & 3 & BNC & V+NP \\
		Gigaword & 17 & 3,964 & -- & -- & -- & 2 & Gigaword & V+NP/PP \\
		IDIX & 52 & 4,022 & 1 & 39 & 540 & 6 & BNC & V+NP/PP \\
		SemEval-2013 & 65 & 4,350 & 4 & 43 & 311 & 4 & ukWaC & unrestricted \\
		\bottomrule
	\end{tabular}
\end{table}

\subsubsection{VNC-Tokens}
The VNC-Tokens dataset contains 53 different PIE types. \citeauthor{Cook2008} extract up to 100 instances from the British National Corpus for each type, for a total of 2,984 instances. These types are based on a pre-existing list of verb-noun combinations and were filtered for frequency and whether two idiom dictionaries both listed them. Instances were extracted automatically, by parsing the corpus and selecting all sentences with the right verb and noun in a direct-object relation. It is unclear whether the extracted sentences were manually checked, but no false extractions are mentioned in the paper or present in the dataset.

All extracted PIE instances were annotated for sense as either \textit{idiomatic}, \textit{literal} or \textit{unclear}. This is a self-explanatory annotation scheme, but \citeauthor{Cook2008} note that senses are not binary, but can form a continuum. For example, the idiomaticity of \textit{have a word} in `You have my word' is different from both the literal sense in `The French have a word for this' and the figurative sense in `My manager asked to have a word'. They instructed annotators to choose \textit{idiomatic} or \textit{literal} even in ambiguous middle-of-the-continuum cases, and restrict the \textit{unclear} label only to cases where there is not enough context to disambiguate the meaning of the PIE.

\subsubsection{Gigaword}
\citet{Sporleder2009} present a corpus of 17 PIE types, for which they extracted all instances from the Gigaword corpus \citep{Gigaword2003}, yielding a total of 3,964 instances. \citeauthor{Sporleder2009} extracted these instances semi-automatically by manually defining all inflectional variants of the verb in the PIE and matching these in the corpus. They did not allow for inflectional variations in non-verb words, nor did they allow intervening words. They annotated these potential idioms as either \textit{literal} or \textit{figurative}, excluding ambiguous and unclear instances from the dataset.

\subsubsection{IDIX}
\label{sec:idix}
\citet{Sporleder2010} build on the methodology of \citet{Sporleder2009}, but annotate a larger set of idioms (52 types) and extract all occurrences from the BNC rather than the Gigaword corpus, for a total of 4,022 instances including false extractions.\footnote{The corpus contains 52 types, rather than the 78/100 types mentioned in the paper, similarly, the actual number of instances in the corpus differs from that reported in the paper. (Caroline Sporleder, personal communication, October 9, 2016)} \citeauthor{Sporleder2010} use a more complex semi-automatic extraction method, which involves parsing the corpus, manually defining the dependency patterns that match the PIE, and extracting all sentences containing those patterns from the corpus. This allows for larger form variations, including intervening words and inflectional variation of all words. In some cases, this yields many non-PIE extractions, as for \textit{recharge one's batteries} in Example~\ref{example:non-pie}. These were not filtered out before annotation, but rather filtered out as part of the annotation process, by having \textit{false extraction} as an additional annotation label. 

For sense annotation, they use an extensive tagset, distinguishing \textit{literal}, \textit{non-literal}, \textit{both}, \textit{meta-linguistic}, \textit{embedded}, and \textit{undecided} labels. Here, the \textit{both} label (Example~\ref{example:both}) is used for cases where both senses are present, often as a form of deliberate word play. The \textit{meta-linguistic} label (Example~\ref{example:metalinguistic}) applies to cases where the PIE instance is used as a linguistic item to discuss, not as part of a sentence. The \textit{embedded} label (Example~\ref{example:embedded}) applies to cases where the PIE is embedded in a larger figurative context, which makes it impossible to say whether a literal or figurative sense is more applicable. The \textit{undecided} label is used for unclear and undecidable cases. They take into account the fact that a PIE can have multiple figurative senses, and enumerate these separately as part of the annotation.

\ex. These high-performance, rugged tools are claimed to offer the best value for money on the market for the enthusiastic d-i-yer and tradesman, and for the first time offer the possibility of a \textbf{battery recharging} time of just a quarter of an hour. (from IDIX corpus, ID \#314)
\label{example:non-pie}

\ex. Left \textbf{holding the baby}, single mothers find it hard to fend for themselves. (from \citealp{Sporleder2010}, p.642)
\label{example:both}

\ex. It has long been recognised that expressions such as to pull someone's leg, to have a bee in one's bonnet, to \textbf{kick the bucket}, to cook someone's goose, to be off one's rocker, round the bend, up the creek, etc. are semantically peculiar. (from \citealp{Sporleder2010}, p.642)
\label{example:metalinguistic}

\ex. You're like a restless bird in a cage. When you get out of the cage, you'll \textbf{fly} very \textbf{high}. (from \citealp{Sporleder2010}, p.642)
\label{example:embedded}

The \textit{both}, \textit{meta-linguistic}, and \textit{embedded} labels are useful and linguistically interesting distinctions, although they occur very rarely (0.69\%, 0.15\%, and an unknown \%, respectively). As such, we include these cases in our tagset (see Section~\ref{sec:corpus-annotation}), but group them under a single label, \textit{other}, to reduce annotation complexity. We also follow \citet{Sporleder2010} in that we combine both the PIE/non-PIE annotation and the sense annotation in a single task.

\subsubsection{SemEval-2013 Task 5b}
\label{sec:semeval-corpus}
\citet{Korkontzelos2013} created a dataset for SemEval-2013 Task 5b, a task on detecting semantic compositionality in context. They selected 65 PIE types from Wiktionary, and extracted instances from the ukWaC corpus \citep{UKWAC}, for a total of 4,350 instances. It is unclear how they extracted the instances, and how much variation was allowed for, although there is some inflectional variation in the dataset. An unspecified amount of manual filtering was done on the extracted instances.

The extracted PIE instances were labelled as \textit{literal}, \textit{idiomatic}, \textit{both}, or \textit{undecidable}. Interestingly, they crowdsourced the sense annotations using CrowdFlower, with high agreement (90\%--94\% pairwise). Undecidable cases and instances on which annotators disagreed were removed from the dataset.

\subsubsection{General Multiword Expression Corpora}
In addition to the aforementioned idiom corpora, there are also corpora focused on multiword expressions (MWEs) in a more general sense. As idioms are a subcategory of MWEs, these corpora also include some idioms. The most important of these are the PARSEME corpus \cite{Savary2018} and the DiMSUM corpus \cite{Schneider2016}. 

DiMSUM provides annotations of over 5,000 MWEs in approximately 90K tokens of English text, consisting of reviews, tweets and TED talks. However, they do not categorise the MWEs into specific types, meaning we cannot easily quantify the number of idioms in the corpus. In contrast to the corpus-specific sense labels seen in other corpora, DiMSUM annotates MWEs with WordNet supersenses, which provide a broad category of meaning for each MWE. 

Similarly, the PARSEME corpus consists of over 62K MWEs in almost 275K tokens of text across 18 different languages (with the notable exception of English). The main differences with DiMSUM, except for scale and multilingualism, are that it only includes verbal MWEs, and that subcategorisation is performed, including a specific category for idioms. Idioms make up almost a quarter of all verbal MWEs in the corpus, although the proportion varies wildly between languages. In both corpora, MWE annotation was done in an unrestricted manner, i.e. there was not a predefined set of expressions to which annotation was restricted.

\subsubsection{Overview}
In sum, there is large variation in corpus creation methods, regarding PIE definition, extraction method, annotation schemes, base corpus, and PIE type inventory. Depending on the goal of the corpus, the amount of deviation that is allowed from the PIE's dictionary form to the instances can be very little \citep{Sporleder2009}, to quite a lot \citep{Sporleder2010}. The number of PIE types covered by each corpus is limited, ranging from 17 to 65 types, often limited to one or more syntactic patterns. The extraction of PIE instances is usually done in a semi-automatic manner, by manually defining patterns in a text or parse tree, and doing some manual filtering afterwards. This works well, but an extension to a large number of PIE types (e.g. several hundreds) would also require a large increase in the amount of manual effort involved. Considering the sense annotations done on the PIE corpora, there is significant variation, with \citet{Cook2008} using only three tags, whereas \citet{Sporleder2010} use six. Outside of PIE-specific corpora there are MWE corpora, which provide a different perspective. A major difference there is that annotation is not restricted to a pre-specified set of expressions, which has not been done for PIEs specifically.

\subsection{Extracting Idioms from Corpora}
\label{sec:prev-work-detection}
There are two main approaches to idiom extraction. The first approach aims to distinguish idioms from other multiword phrases, where the main purpose is to expand idiom inventories with rare or novel expressions
\citep[for example]{Fazly2009,Muzny2013,Gong2016,Senaldi2016}.
The second approach aims to extract all occurrences of a known idiomatic expression in a text. In this paper, we focus on the latter approach. We rely on idiom dictionaries to provide a list of PIE types, and build a system that extracts all instances of those PIE types from a corpus. High-quality idiom dictionaries exist for most well-resourced languages, but their reliability and coverage is not known. As such, we quantify the coverage of dictionaries in Section~\ref{sec:dictionary-comparison}. 

There is, to the best of our knowledge, no existing work that focuses on dictionary-based PIE extraction. However, there is closely-related work by \citet{Inurrieta2016}, who present a system for the dictionary-based extraction of verb-noun combinations (VNCs) in English and Spanish. In their case, the VNCs can be any kind of multiword expression, which they subdivide into literal expressions, collocations, light verb constructions, metaphoric expressions, and idioms. They extract 173 English VNCs and 150 Spanish VNCs and annotate these with both their lexico-semantic MWE type and the amount of morphosyntactic variation they exhibit. \citeauthor{Inurrieta2016} then compare a word sequence-based method, a chunking-based method, and a parse-based method for VNC extraction. Each method relies on the morpho-syntactic information in order to limit false extractions. Precision is evaluated manually on a sample of the extracted VNCs, and recall is estimated by calculating the overlap between the output of the three methods. Evaluation shows that the methods are highly complementary both in recall, since they extract different VNCs, and in precision, since combining the extractors yields fewer false extractions. 

Whereas \citet{Inurrieta2016} focus on both idiomatic and literal uses of the set of expressions, like in this paper, \citet{Savary2017a} tackle only half of that task, namely extracting only literal uses of a given set of VMWEs in Polish. This complicates the task, since it combines extracting all occurrences of the VMWEs and then distinguishing literal from idiomatic uses. Interestingly, they also experiment with models of varying complexity, i.e. just words, part-of-speech tags, and syntactic structures. Their results are hard to put into perspective however, since the frequency of literal VMWEs in their corpus is very rare, whereas corpora containing PIEs tend to show a more balanced distribution.

Other similar work to ours also focuses on MWEs more generally, or on different subtypes of MWEs. In addition, these tend to combine both extraction and disambiguation in that they aim to extract only idiomatically used instances of the MWE, without extracting literally used instances or non-instances. Within this line of work, \citet{Baldwin2005} focuses on verb-particle constructions, \citet{Boukobza2009} on verbal MWEs (including idioms), and \citet{Pasquer2018} on verbal MWEs (especially non-canonical variants). 

Both \citeauthor{Boukobza2009} and \citeauthor{Pasquer2018} rely on a pre-defined set of expressions, whereas \citeauthor{Baldwin2005} also extracts unseen expressions, although based on a pre-defined set of particles and within the vary narrow syntactic frame of verb-particle constructions. The work of \citeauthor{Baldwin2005} is most similar to ours in that it builds an unsupervised system using existing NLP tools (PoS taggers, chunkers, parsers) and finds that a combination of systems using those tools performs best, as we find in Section~\ref{sec:extraction-results}. \citeauthor{Boukobza2009} and \citeauthor{Pasquer2018}, by contrast, use supervised classifiers which require training data, not just for the task in general, but specific to the set of expressions used in the task.

Although our approach is similar to that of \citeauthor{Inurrieta2016}, both in the range of methods used and in the goal of extracting certain multiword expressions regardless of morphosyntactic variation, there are two main differences. First, we use dictionaries, but extract entries automatically and do not manually annotate their type and variability. As a result, our methods rely only on the surface form of the expression taken from the dictionary. Second, we evaluate precision and recall in a more rigorous way, by using an evaluation corpus exhaustively annotated for PIEs. In addition, we do not put any restriction on the syntactic type of the expressions to be extracted, which \citet{Baldwin2005}, \citet{Boukobza2009}, \citet{Inurrieta2016}, and \citet{Pasquer2018} all do. 


\section{Coverage of Idiom Inventories}
\label{sec:dictionary-comparison}

\subsection{Background}
\label{sec:evaluation}
Since our goal is developing a dictionary-based system for extracting potentially idiomatic expressions, we need to devise a proper method for evaluating such a system. This is not straightforward, even though
the final goal of such a system is simple: it should extract all potentially idiomatic expressions from a corpus and nothing else, regardless of their sense and the form they are used in. 
The type of system proposed here hence has two aspects that can be evaluated: the dictionary that it is using as a resource for idiomatic expression, and the extractor component that finds idioms in a corpus.

The difficulty here is that there is no undisputed and unambiguous definition of what counts as an idiom \cite{Nunberg1994}, as is the case with multiword expressions in general \cite{Constant2017}. Of course, a complete set of idiomatic expressions for English (or any other language) is impossible to get due to the broad and ever-changing nature of language. This incompleteness is exacerbated by the ambiguity problem: if we had a clear definition of idiom we could make an attempt of evaluating idiom dictionaries on their accuracy, but it is practically impossible to come up with a definition of idiom that leaves no room for ambiguity. \footnote{For example, one problem is posed by the main characteristic of idioms, \textit{semantic non-compositionality}. When judging this characteristic, it is often unclear whether the meaning of an idiomatic expression is a non-compositional combination of the literal senses of its component words, or a compositional combination of figurative senses of the component words. Consider, for example, the expression \textit{cheap and nasty}, `of low cost and bad quality'. If one sees `bad quality' as being one of the senses of \textit{nasty}, this could be considered compositional. If not, however, the meaning of the idiom would have to arise from non-compositional combination.}
This ambiguity, among others, creates a large grey area between clearly non-idiomatic phrases on the one hand (e.g. \textit{buy a house}), and clear potentially idiomatic phrases on the other hand (e.g. \textit{buy the farm}). 
As a consequence, we cannot empirically evaluate the coverage of the dictionaries. Instead, in this work, we will quantify the divergence between various idiom dictionaries and corpora, with regard to their idiom inventories. If they show large discrepancies, we take that to mean that either there is little agreement on definitions of idiom or the category is so broad that a single resource can only cover a small proportion. Conversely, if there is large agreement, we assume that idiom resources are largely reliable, and that there is consensus around what is, and what is not, an idiomatic expression.

We use different idiom resources and assume that the combined set of resources yields an approximation of the true set of idioms in English. A large divergence between the idiom inventories of these resources would then suggest a low recall for a single resource, since many other idioms are present in the other resources. Conversely, if the idiom inventories largely overlap, that indicates that a single resource can already yield decent coverage of idioms in the English language. The results of the dictionary comparisons are in Section~\ref{sec:dictionary-comparison-results}.

\subsection{Selected Idiom Resources (Data and Method)}
We evaluate the quality of three idiom dictionaries by comparing them to each other and to three idiom corpora. Before we report on the comparison we first describe why we select and how we prepare these resources. 
We investigate the following six idiom resources: 

\begin{enumerate}
   \item Wiktionary\footnote{http://en.wiktionary.org};
   \item the Oxford Dictionary of English Idioms (ODEI, \citealp{Ayto2009}); 
   \item UsingEnglish.com (UE)\footnote{http://www.usingenglish.com/reference/idioms};
    \item the Sporleder corpus \citep{Sporleder2010};
    \item the VNC dataset \citep{Cook2008};
    \item and the SemEval-2013 Task 5 dataset \citep{Korkontzelos2013}. 
\end{enumerate}

These dictionaries were selected because they are available in digital format. Wiktionary and UsingEnglish have the added benefit of being freely available. However, they are both crowdsourced, which means they lack professional editing. In contrast, ODEI is a traditional dictionary, created and edited by lexicographers, but it has the downside of not being freely available.

For Wiktionary, we extracted all idioms from the category `English Idioms'\footnote{http://en.wiktionary.org/wiki/Category:English\_idioms} from the English version of Wiktionary. We took the titles of all pages containing a dictionary entry and considered these idioms. Since we focus on multiword idiomatic expressions, we filtered out all single-word entries in this category. More specifically, since Wiktionary is a constantly changing resource, we used the 8,482 idioms retrieved on 10-03-2017, 15:30. 
We used a similar extraction method for UE, a web page containing freely available resources for ESL learners, including a list of idioms. We extracted all idioms which have publicly available definitions, which numbered 3,727 on 10-03-2017, 15:30. Again, single-word entries and duplicates were filtered out. %
Concerning ODEI, all idioms from the e-book version were extracted, amounting to 5,911 idioms scraped on 13-03-2017, 10:34. Here we performed an extra processing step to expand idioms containing content in parentheses, such as \textit{a tough (or hard) nut (to crack)}. Using a set of simple expansion rules and some hand-crafted exceptions, we automatically generated all variants for this idiom, with good, but not perfect accuracy. For the example above, the generated variants are: \textit{\{a tough nut, a tough nut to crack, a hard nut, a hard nut to crack\}}. The idioms in the VNC dataset are in the form \texttt{verb\_noun}, e.g. \texttt{blow\_top}, so they were manually expanded to a regular dictionary form, e.g. \textit{blow one's top} before comparison.

\subsection{Method}
\label{sec:dictionary-comparison-method}
In many cases, using simple string-match to check overlap in idioms does not work, as exact comparison of idioms misses equivalent idioms that differ only slightly in dictionary form. Differences between resources are caused by, for example:

\begin{itemize}
\item inflectional variation (\textit{crossing the Rubicon} --- \textit{cross the Rubicon});
\item variation in scope (\textit{as easy as ABC} --- \textit{easy as ABC});
\item determiner variation (\textit{put the damper on} --- \textit{put a damper on});
\item spelling variation (\textit{mind your p's and q's} --- \textit{mind your ps and qs});
\item order variation (\textit{call off the dogs} --- \textit{call the dogs off});
\item and different conventions for placeholder words (\textit{recharge your batteries} --- \textit{recharge one's batteries}), where both \textit{your} and \textit{one's} can generalise to any possessive personal pronoun. 
\end{itemize}

These minor variations do not fundamentally change the nature of the idiom, and we should count these types of variation as belonging to the same idiom (see also \citealp{Pasquer2018b}, who devise a measure to quantify different types of variation allowed by specific MWEs). So, to get a good estimate of the true overlap between idiom resources, these variations need to be accounted for, which we do in our flexible matching approach. 

There is one other case of variation not listed above, namely lexical variation (e.g. \textit{rub someone up the wrong way} - \textit{stroke someone the wrong way}). We do not abstract over this, since we consider lexical variation to be a more fundamental change to the nature of the idiom. That is, a lexical variant is an indicator of the coverage of the dictionary, where the other variations are due to different stylistic conventions and do not indicate actual coverage. In addition, it is easy to abstract over the other types of variation in an NLP application, but this is not the case for lexical variation.

The overlap counts are estimated by abstracting over all variations except lexical variation in a semi-automatic manner, using heuristics and manual checking. Potentially overlapping idioms are selected using the following set of heuristics: whether an idiom from one resource is a substring (including gaps) of an idiom in the other resource, whether the words of an idiom form a subset of the words of an idiom in the other resource, and whether there is an idiom in the other resource which has a Levenshtein ratio%
\footnote{As computed by \texttt{ratio()} from the \texttt{python-Levenshtein} package.}
of over 0.8. The Levenshtein ratio is an indicator of the Levenshtein distance between the two idioms relative to their length. These potential matches are then judged manually on whether they are really forms of the same idiom or not. 

\subsection{Results}
\label{sec:dictionary-comparison-results}
\begin{figure}[bt]
	\centering
	\includegraphics[width=0.92\textwidth]{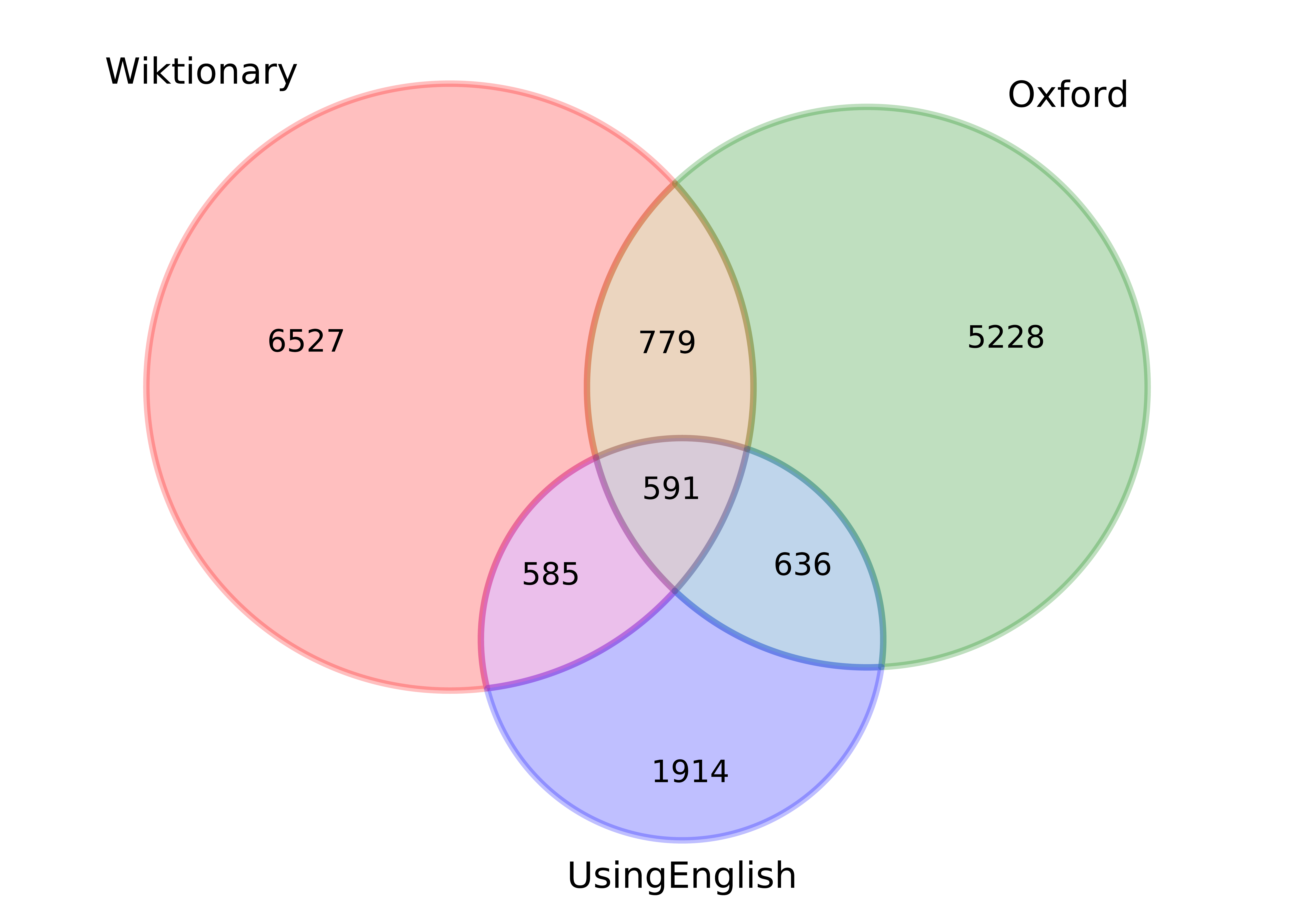}
	\caption{Venn diagram of case-insensitive exact string match overlap between the three idiom dictionaries. Note that the numbers in this figure are based on exact string matching, so they differ from the numbers in Table~\ref{table:idiom-comparison-partial}, matching of similar, but not identical idioms, as described in Section~\ref{sec:dictionary-comparison-method}.}
	\label{figure:dictionary-overlap-venn}
\end{figure}

The results of using exact string matching to quantify the overlap between the dictionaries is illustrated in Figure~\ref{figure:dictionary-overlap-venn}.

Overlap between the three dictionaries is low. A possible explanation for this lies with the different nature of the dictionaries. Oxford is a traditional dictionary, created and edited by professional lexicographers, whereas Wiktionary is a crowdsourced dictionary open to everyone, and UsingEnglish is similar, but focused on ESL-learners. It is likely that these different origins result in different idiom inventories. Similarly, we would expect that the overlap between a pair of traditional dictionaries, such as the ODEI and the Penguin Dictionary of English Idioms \citep{Gulland2002} would be significantly higher. It should also be noted, however, that comparisons between more similar dictionaries also found relatively little overlap (\citealp{Ide1994}; \citealp[p.99]{Seret2008}). A counterpoint is provided by \citet{Villavicencio2005}, who quantifies coverage of verb-particle constructions in three different dictionaries and finds large overlap -- perhaps because verb-particle are a more restricted class.

As noted previously, using exact string matching is a very limited approach to calculating overlap. Therefore, we used heuristics and manual checking to get more precise numbers, as shown in Table~\ref{table:idiom-comparison-partial}, which also includes the three corpora in addition to the three dictionaries. As the manual checking only involved judging similar idioms found in pairs of resources, we cannot calculate three-way overlap as in Figure~\ref{figure:dictionary-overlap-venn}. The counts of the pair-wise overlap between dictionaries differ significantly between the two methods, which serves to illustrate the limitations of using only exact string matching and the necessity of using more advanced methods and manual effort.

Several insights can be gained from the data in Table~\ref{table:idiom-comparison-partial}. The relation between Wiktionary and the SemEval corpus is obvious (cf. Section~\ref{sec:semeval-corpus}), given the 96.92\% coverage.\footnote{One would expect 100\% coverage here, but Wiktionary is an ever-changing resource and has changed since the creation of the SemEval corpus.} For the other dictionary-corpus pairs, the coverage increases proportionally with the size of the dictionary, except in the case of UsingEnglish and the Sporleder corpus. The proportional increase indicates no clear qualitative differences between the dictionaries, i.e. one does not have a significantly higher percentage of non-idioms than the other, when compared to the corpora.

\begin{table}
	\centering
	\caption{Percentage of overlapping idioms between different idiom resources, abstracting over minor variations. Value is the number of idioms in the intersection of two idiom sets, as a percentage of number of idioms in the resource in the left column. For example, 56.60\% of idioms in the VNC occur in Wiktionary.}
	\label{table:idiom-comparison-partial}
	\setlength{\tabcolsep}{5.5pt}
	\begin{tabular}{l||c|c|c|c|c|c}
		\toprule
		& \textbf{Wiktionary} & \textbf{ODEI} & \textbf{UE} & \textbf{Sporleder} & \textbf{VNC} & \textbf{SemEval} \\ \midrule
		\textbf{Wiktionary} & 100\% & 28.99\% & 20.60\% & 0.38\% & 0.40\% & 0.87\% \\ \midrule
		\textbf{ODEI} & 34.12\% & 100\% & 29.22\% & 0.46\% & 0.36\% & 0.69\% \\ \midrule
		\textbf{UE} & 44.73\% & 54.57\% & 100\% & 0.94\% & 0.54\% & 0.99\% \\ \midrule
		\textbf{Sporleder} & 60.78\% & 60.78\% & 68.63\% & 100\% & 3.92\% & 3.92\% \\ \midrule
		\textbf{VNC} & 56.60\% & 45.28\% & 35.85\% & 3.77\% & 100\% & 1.89\% \\ \midrule
		\textbf{SemEval} & 96.92\% & 70.77\% & 52.31\% & 3.08\% & 1.54\% & 100\% \\ \bottomrule
	\end{tabular}
\end{table}

Generally, overlap between dictionaries and corpora is low: the two biggest, ODEI and Wiktio\-na\-ry have only around 30\% overlap, while the dictionaries also cover no more than approximately 70\% of the idioms used in the various corpora. Overlap between the three corpora is also extremely low, at below 5\%. This is unsurprising, since a new dataset is more interesting and useful when it covers a different set of idioms than used in an existing dataset, and thus is likely constructed with this goal in mind. 

\section{Corpus Annotation}
\label{sec:corpus-annotation}
In order to evaluate the PIE extraction methods developed in this work (Section~\ref{sec:pie-extraction}), we exhaustively annotate an evaluation corpus with all instances of a pre-defined set of PIEs. As part of this, we come up with a workable definition of PIEs, and measure the reliability of PIE annotation by inter-annotator agreement.

Assuming that we have a set of idioms, the main problem of defining what is and what is not a potentially idiomatic expression is caused by variation. In principle, potentially idiomatic expression is an instance of a phrase that, when seen without context, could have either an idiomatic or a literal meaning. This is clearest for the dictionary form of the idiom, as in Example~\ref{example:pie-1}. Literal uses generally allow all kinds of variation, but not all of these variations allow a figurative interpretation, e.g. Example~\ref{example:pie-2}. However, how much variation an idiom can undergo while retaining its figurative interpretation is different for each expression, and judgements of this might vary from one speaker to the other. An example of this is \textit{spill the bean}, a variant of \textit{spill the beans}, in Example~\ref{example:pie-3} judged by \citet[p.65]{Fazly2009} as being highly questionable. However, even here a corpus example can be found containing the same variant used in a figurative sense (Example~\ref{example:pie-4}). 

As such, we assume that we cannot know a priori which variants of an expression allow a figurative reading, and are thus a potentially idiomatic expression. Therefore we consider every possible morpho-syntactic variation of an idiom a PIE, regardless of whether it actually allows a figurative reading. We believe the boundaries of this variation can only be determined based on corpus evidence, and even then they are likely variable. 

Note that a similar question is tackled by \citet{Savary2017a}, when they establish the boundary between a `literal reading of a VMWE' and a `coincidental co-occurrence'. \citeauthor{Savary2017a}'s answer is similar to ours, in that they count something as a literal reading of a VMWE if it `the same or equivalent dependencies hold between [the expression]'s components as in its canonical form'.

\ex. John \textbf{kicked the bucket} last night.
\label{example:pie-1}

\ex. * \textbf{The bucket}, John \textbf{kicked} last night.
\label{example:pie-2}

\ex. ?? Azin \textbf{spilled the bean}. (from \citealp[p.65]{Fazly2009})
\label{example:pie-3}

\ex. Alba reveals Fantastic Four 2 details The Invisible Woman actress \textbf{spills the bean} on super sequel (from ukWaC)
\label{example:pie-4}

\subsection{Evaluating the Extraction Methods}
\label{sec:evaluation-of-extraction}
Evaluating the extraction methods is easier than evaluating dictionary coverage, since the goal of the extraction component is more clearly delimited: given a set of PIEs from one or more dictionaries, extract all occurrences of those PIEs from a corpus. Thus, rather than dealing with the undefined set of all PIEs, we can work with a clearly defined and finite set of PIEs from a dictionary.

Because we have a clearly defined set of PIEs, we can exhaustively annotate a corpus for PIEs, and use that annotated corpus for automatic evaluation of extraction methods using recall and precision. This allows us to facilitate and speed up annotation by pre-extracting sentences possibly containing a PIE. After the corpus is annotated, the precision and recall can be easily estimated by comparing the extracted PIE instances to those marked in the corpus. The details of the corpus selection, dictionary selection, extraction heuristic and annotation procedure are presented in Section \ref{sec:annotation-procedure}, and the details and results of the various extraction methods are presented in Section~\ref{sec:pie-extraction}.

\subsection{Base Corpus and Idiom Selection}
As a base corpus, we use the XML version of the British National Corpus \citep{BNC-XML}, because of its size, variety, and wide availability.\footnote{The BNC is freely available under the BNC User Licence at http://ota.ox.ac.uk/desc/2554.} The BNC is pre-segmented into \textit{s-units}, which we take to be sentences, \textit{w-units}, which we take to be words, and \textit{c-units}, punctuation. We then extract the text of all w-units and c-units. We keep the sentence segmentation, resulting in a set of plain text sentences. All sentences are included, except for sentences containing \verb|<gap>| elements, which are filtered out. These \verb|<gap>| elements indicate places where material from the original has been left out, e.g. for anonymisation purposes. Since this can result in incomplete sentences that cannot be parsed correctly, we filter out sentences containing these gaps. 

We use only the written part of the BNC. From this, we extract a set of documents with the aim of having as much genre variation as possible. To achieve this, we select the first document in each genre, as defined by the \verb|classCode| attribute (e.g. \verb|nonAc|, \verb|commerce|, \verb|letters|). The resulting set of 46 documents makes up our base corpus. Note that these documents vary greatly in size, which means the resulting corpus is varied, but not balanced in terms of size (Table~\ref{table:base-corpus-size}). The documents are split across a development and test set, as specified at the end of Section~\ref{sec:annotation-procedure}. We exclude documents with IDs starting with \verb|A0| from all annotation and evaluation procedures, as these were used during development of the extraction tool and annotation guidelines.

\begin{table}[htbp]
    \caption{Statistics on the size of the BNC documents used for PIE annotation and the split in development and test set.}
    \centering
    \begin{tabular}{lrrrr}
        \toprule
         & Documents & Tokens & Shortest Document & Longest Document \\ 
         \midrule
         Development & 22 & 832,083 & 1,815 & 228,230 \\
         Test & 23 & 814,125 & 1,984 & 231,846 \\
         \midrule
         Total & 45 & 1,646,208 & 1,815 & 231,846 \\
         \bottomrule
    \end{tabular}
    \label{table:base-corpus-size}
\end{table}

As for the set of potentially idiomatic expressions, we use the intersection of the three dictionaries, Wiktionary, Oxford, and UsingEnglish. Based on the assumption that, if all three resources include a certain idiom, it must unquestionably be an idiom, we choose the intersection (also see Figure~\ref{figure:dictionary-overlap-venn}). This serves to exclude questionable entries, like \textit{at all}, which is in Wiktionary. The final set of idioms used for these experiments consists of 591 different multiword expressions. Although we aim for wide coverage, this is a necessary trade-off to ensure quality. At the same time, it leaves us with a set of idiom types that is approximately ten times larger than present in existing corpora. The set of 591 idioms includes idioms with a large variety of syntactic patterns, of which the most frequent ones are shown in Table~\ref{table:syntactic-patterns}. The statistics show that the types most prevalent in existing corpora, verb-noun and preposition-noun combinations, are indeed the most frequent ones, but that there is a sizeable minority of types that do not fall into those categories, including coordinated adjectives, coordinated nouns, and nouns with prepositional phrases. This serves to emphasise the necessity of not restricting corpora to a small set of syntactic patterns.

\begin{table}[hbtp]
    \centering
    \caption{The 10 most frequent syntactic patterns in the set of 591 idiomatic expressions, based on automatic part-of-speech tags produced by Spacy which were manually corrected.}
    \label{table:syntactic-patterns}
    \begin{tabular}{llr}
        \toprule
        Pattern & Example & Count \\
        \midrule
        VERB-DET-NOUN & \textit{take the plunge} & 92 \\
        ADP-DET-NOUN & \textit{in a nutshell} & 59 \\
        VERB-NOUN & \textit{pull rank} & 40 \\
        VERB-ADP-DET-NOUN & \textit{smoke like a chimney} & 22 \\
        ADJ-NOUN & \textit{small potatoes} & 21 \\
        NOUN-CCONJ-NOUN & \textit{smoke and mirrors} & 17 \\
        NOUN-ADP-NOUN & \textit{word of mouth} & 15 \\
        ADJ-ADP-DET-NOUN & \textit{rough around the edges} & 13 \\
        ADP-NOUN & \textit{on ice} & 12 \\
        ADJ-CCONJ-ADJ & \textit{black and blue} & 12 \\
        VERB-ADP-NOUN & \textit{play with fire} & 11 \\
        NOUN-ADP-DET-NOUN & \textit{hair of the dog} & 10 \\
        \bottomrule
    \end{tabular}
\end{table}

\subsection{Extraction of PIE Candidates}
\label{sec:candidate-extraction}
To annotate the corpus completely manually would require annotators to read the whole corpus, and cross-reference each sentence to a list of almost 600 PIEs, to check whether one of those PIEs occurs in a sentence. We do not consider this a feasible annotation settings, due to both the difficulty of recognising literal usages of idioms and the time cost needed to find enough PIEs, given their low overall frequency. As such, we use a pre-extraction step to present candidates for annotation to the human annotators. 

Given the corpus and the set of PIEs, we heuristically extract the PIE candidates as follows: given an idiomatic expression, extract every sentence which contains all the defining words of the idiom, in any form. This ensures that all possibly matching sentences get extracted, while greatly pruning the amount of sentences for annotators to look at. In addition, it allows us to present the heuristically matched PIE type and corresponding words to the annotators, which makes it much easier to judge whether something is a PIE or not. This also means that annotators never have to go through the full list of PIEs during the annotation process. 

Initially, the heuristic simply extracted any sentence containing all the required words, where a word is any of the inflectional variants of the words in the PIE, except for determiners and punctuation. This method produced large amounts of noise, that is, a set of PIE candidates with only a very low percentage of actual PIEs. This was caused by the presence of some highly frequent PIEs with very little defining lexical content, such as \textit{on the make}, and \textit{in the running}. For example, with the original method, every sentence containing the preposition \textit{on}, and any inflectional form of the verb \textit{make} was extracted, resulting in a huge number of non-PIE candidates.

To limit the amount of noise, two restrictions were imposed. The first restrictions disallows word order variation for PIEs which do not contain a verb. The rationale behind this is that word order variation is only possible with PIEs like \textit{spill the beans} (e.g. \textit{the beans were spilled}), and not with PIEs like \textit{in the running} (\textit{*the running in??}). The second restriction is that we limit the number of words that can be inserted between the words of a PIE, but only for PIEs like \textit{on the make}, and \textit{in the running}, i.e. PIEs which only contain prepositions, determiners and a single noun. The number of intervening words was limited to three tokens, allowing for some variation, as in Example~\ref{ex:acceptable-variation}, but preventing sentences like Example~\ref{ex:excessive-variation} from being extracted. This restriction could result in the loss of some PIE candidates with a large number of intervening words. However, the savings in annotation time clearly outweigh the small loss in recall in this situation.

\ex. Either at New Year or before July you can anticipate a change \textbf{in} \textbf{the} everyday \textbf{running} of your life. (\textit{in the running} - BNC - document CBC - sentence 458)
\label{ex:acceptable-variation}

\ex. [..] if [he] hung around near the goal or \textbf{in} \textbf{the} box for that matter instead of \textbf{running} all over the show [..] (\textit{in the running} - BNC - document J1C - sentence 1341)
\label{ex:excessive-variation}

\subsection{Annotation Procedure}
\label{sec:annotation-procedure}
The manual annotation procedure consists of three different phases (pilot, double annotation, single annotation), followed by an adjudication step to resolve conflicting annotations. Two things are annotated: whether something is a PIE or not, and if it is a PIE, which sense the PIE is used in. In the first phase (\texttt{0-100-*}), we randomly select hundred of the 2,239 PIE candidates which are then annotated by three annotators. All annotators have a good command of English, are computational linguists, and familiar with the subject. The annotators include the first and last author of this paper. 

The annotators were provided with a short set of guidelines, of which the main rule-of-thumb for labelling a phrase as a PIE is as follows: any phrase is a PIE when it contains all the words, with the same part-of-speech, and in the same grammatical relations as in the dictionary form of the PIE, ignoring determiners.\footnote{Note that, while not exactly the same relation, we do allow for passivisation, e.g. `The trick was done by using a new approach' for \textit{do the trick}. For the full guidelines, see the repository at https://github.com/hslh/pie-detection.} 

For sense annotation, annotators were to mark a PIE as idiomatic if it had a sense listed in one of the idiom dictionaries, and as literal if it had a meaning that is a regular composition of its component words. For cases which were undecidable due to lack of context, the \textit{?}-label was used. The \textit{other}-label was used as a container label for all cases in which neither the literal or idiomatic sense was correct (e.g. meta-linguistic uses and embeddings in metaphorical frames, see also Section~\ref{sec:idix}).

The first phase of annotation serves to bring to light any inconsistencies between annotators and fill in any gaps in the annotation guidelines. The resulting annotations already show a reasonably high agreement of 0.74 Fleiss' Kappa. Table~\ref{table:annotation-statistics} shows annotation details and agreement statistics for all three phases. The annotation tasks suffixed by \verb|-PIE| indicate agreement on PIE/non-PIE annotation and the tasks suffixed by \verb|-sense| indicate agreement on sense annotation for PIEs.

\begin{table}[hbtp]
	\centering
	\caption{Details of the annotation phases and inter-annotator agreement statistics. The number of candidates for sense annotation is the number on which all annotators initially agreed that it was a PIE, i.e. pre-adjudication. Note that sense and PIE annotation are split here for clarity of presentation; in practice they were annotated as a joint task.}
	\label{table:annotation-statistics}
	\begin{tabular}{lrrrr}
		\toprule 
		Annotation Task & \# Annotators & \# Candidates & \% Agreement & Fleiss' Kappa \\ 
		\midrule
		0-100-PIE & 3 & 100 & 0.87 & 0.74 \\
		100-600-PIE & 2 & 500 & 0.96 & 0.91 \\
		600-1100-PIE & 2 & 500 & 0.94 & 0.88 \\
		1100-2239-PIE & 1 & 1139 & n/a & n/a \\
		\midrule
		0-100-sense & 3 & 38 & 0.82 & 0.65 \\
		100-600-sense & 2 & 160 & 0.92 & 0.83 \\
		600-1100-sense & 2 & 259 & 0.79 & 0.63 \\
		1100-2239-sense & 1 & 558 & n/a & n/a \\
		\bottomrule
	\end{tabular}
\end{table}

In the second phase of annotation (\texttt{100-600-* \& 600-1100-*}), another 1000 of the 2239 PIE candidates are selected to be annotated by two pairs of annotators. This shows very high agreement, as shown in Table~\ref{table:annotation-statistics}. This is probably due to the improvement in guidelines and the discussion following the pilot round of annotation. The exception to this are the somewhat lower scores for the \verb|600-1100-sense| annotation task. Adjudication revealed that this is due almost exclusively because of a different interpretation of the literal and idiomatic senses of a single PIE type: \textit{on the ground}. Excluding this PIE type, Fleiss' Kappa increases from 0.63 to 0.77.

Because of the high agreement on PIE annotation, we deem it sufficient for the remainder (1108 candidates) to be annotated by only the primary annotator in the third phase of annotation (\texttt{1100-2239-*}).  The reliability of the single annotation can be checked by comparing the distribution of labels to the multi-annotated parts. This shows that it falls clearly within the ranges of the other parts, both in the proportion of PIEs and idiomatic senses (see Table~\ref{table:distributional-statistics}). The single-annotated part has 49.0\% PIEs, which is only 4 percentage points above the 44.7\% PIEs in the multi-annotated parts. The proportion of idioms is just 2 percentage points higher, with 55.9\% versus 53.9.\%.

\begin{table}[hbtp]
	\centering
	\caption{Distributional statistics on the annotated PIE corpus, post-adjudication. Adjudication resolved all instances which were disagreed upon and all \textit{?}-sense-labels, hence the presence of only 3 sense labels: i(diomatic), l(iteral), and o(ther).}
	\label{table:distributional-statistics}
	\begin{tabular}{lrrrrr}
		\toprule 
		Part & \# Candidates & PIE (y/n) & \% PIE & sense (i/l/o) & \% Idiom \\ \midrule
		0-100 & 100 & 47/53 & 47.0 & 23/24/0 & 48.9 \\
		100-600 & 500 & 169/331 & 33.4 & 112/54/3 & 66.3 \\
		600-1100 & 500 & 276/224 & 55.2 & 130/132/14 & 47.1 \\
		1100-2239 & 1139 & 558/581 & 49.0 & 312/229/17 & 55.9 \\
		\midrule
		Total & 2239 & 1050/1189 & 46.9 & 577/439/34 & 55.0  \\
		\bottomrule
	\end{tabular}
\end{table}

Although inter-annotator agreement was high, there was still a significant number of cases in the triple and double annotated PIE candidate sets where not all annotators agreed. These cases were adjudicated through discussion by all annotators, until they were in agreement. In addition, all PIE candidates which initially received the \textit{?}-label (unclear or undecidable) for sense or PIE were resolved in the same manner. In the adjudication procedure, annotators were provided with additional context on each side of the idiom, in contrast to the single sentence provided during the initial annotation. The main reason to do adjudication, rather than simply discarding all candidates for which there was disagreement, was that we expected exactly those cases for which there are conflicting annotations to be the most interesting ones, since having non-standard properties would cause the annotations to diverge. Examples of such interesting non-standard cases are \textit{at sea} as part of a larger satirical frame in Example~\ref{ex:satire} and \textit{cut the mustard} in Example~\ref{ex:headline} where it is used in a headline as wordplay on a Cluedo character.

\ex. The bovine heroine has connections with Cowpeace International, and deals with a huge treacle slick \textbf{at sea}. (\textit{at sea} - BNC - document CBC - sentence 13550)
\label{ex:satire}

\ex. Why not \textbf{cut the Mustard}? [..] WADDINGTON Games's proposal to axe Reverend Green from the board game Cluedo is a bad one. (\textit{cut the mustard} - BNC - document CBC - sentence 14548)
\label{ex:headline}

We split the corpus at the document level. The corpus consists of 45 documents from the BNC, and we split it in such a way that the development set has 1,112 candidates across 22 documents and the test set has 1,127 candidates from 23 documents. Note that this means that the development and test set contain different genres. This ensures that we do not optimise our systems on genre-specific aspects of the data.


\section{Dictionary-based PIE Extraction}
\label{sec:pie-extraction}
We propose and implement four different extraction methods, of differing complexities: exact string match, fuzzy string match, inflectional string match, and parser-based extraction. Because of the absence of existing work on this task, we compare these methods to each other, where the more basic methods function as baselines. More complex methods serve to shine light on the difficulty of the PIE extraction task; if simple methods already work sufficiently well, the task is not as hard as expected, and vice versa. Below, each of the extraction methods is presented and discussed in detail. 

\subsection{String-based Extraction Methods}
\paragraph{Exact String Match} This is, very simply, extracting all instances of the exact dictionary form of the PIE, from the tokenized text of the corpus. Word boundaries are taken into account, so \textit{at sea} does not match `th\textbf{at sea}water'. As a result, all inflectional and other variants of the PIE are ignored.

\paragraph{Fuzzy String Match} Fuzzy string match is a rough way of dealing with morphological inflection of the words in a PIE. We match all words in the PIE, taking into account word boundaries, and allow for up to 3 additional letters at the end of each word. These 3 additional characters serve to cover inflectional suffixes. 

\paragraph{Inflectional String Match} In inflectional string match, we aim to match all inflected variations of a PIE. This is done by generating all morphological variants of the words in a PIE, generating all combinations of those words, and then using exact string match as described earlier. 

Generating morphological variations consists of three steps: part-of-speech tagging, morphological analysis, and morphological reinflection. Since inflectional variation only applies to verbs and nouns, we use the Spacy\footnote{http://spacy.io} part-of-speech tagger to detect the verbs and nouns. Then, we apply the morphological analyser \verb|morpha| to get the base, uninflected form of the word, and then use the morphological generation tool \verb|morphg| to get all possible inflections of the word. Both tools are part of the Morph morphological processing suite \citep{Minnen2001}. Note that the Morph tools depend on the part-of-speech tag in the input, so that a wrong PoS may lead to an incorrect set of morphological variants.

For a PIE like \textit{spill the beans}, this results in the following set of variants: $\{$\textit{spill the bean, spills the bean, spilled the bean, spilling the bean, spill the beans, spills the beans, spilled the beans, spilling the beans}$\}$. Since we generate up to 2 variants for each noun, and up to 4 variants for each verb, the number of variants for PIEs containing multiple verbs and nouns can get quite large. On average, 8 additional variants are generated for each potentially idiomatic expression.

\paragraph{Additional Steps} 
For all string match-based methods, ways to improve performance are implemented, to make them as competitive as possible. Rather than doing exact string matching, we also allow words to be separated by something other than spaces, e.g. \textit{nuts-and-bolts} for \textit{nuts and bolts}. Additionally, there is an option to take into account case distinctions. With the \textit{case-sensitive} option, case is preserved in the idiom lists, e.g. \textit{coals to Newcastle}, and the string matching is done in a case-sensitive manner. This increases precision, e.g. by avoiding PIEs as part of proper names, but also comes at a cost of recall, e.g. for sentence-initial PIEs. Thirdly, there is the option to allow for a certain number of intervening words between each pair of words in the PIE. This should improve recall, at the cost of precision. For example, this would yield the true positive \textit{make a huge mountain out of a molehill} for \textit{make a mountain out of a molehill}, but also false positives like \textit{have a smoke and go} for \textit{have a go}.

A third shared property of the string-based methods is the processing of placeholders in PIEs. PIEs containing possessive pronoun placeholders, such as \textit{one's} and \textit{someone's} are expanded. That is, we remove the original PIE, and add copies of the PIE where the placeholder is replaced by one of the possessive personal pronouns. For example, \textit{a thorn in someone's side} is replaced by \textit{a thorn in \{my, your, his, ...\} side}. In the case of \textit{someone's}, we also add a wildcard for any possessively used word, i.e. \textit{a thorn in \textemdash's side}, to match e.g. \textit{a thorn in Google's side}. Similarly, we make sure that PIE entries containing \textit{\textemdash}, such as \textit{the mother of all \textemdash}, will match any word for \textit{\textemdash} during extraction. We do the same for \textit{someone}, for which we substitute objective pronouns. For \textit{one}, this is not possible, since it is too hard to distinguish from the \textit{one} used as a number. 

\subsection{Parser-Based Extraction Methods} 
Parser-based extraction is potentially the widest-coverage extraction method, with the capacity to extract both morphological and syntactic variants of the PIE. This should be robust against the most common modifications of the PIE, e.g. through word insertions (\textit{spill all the beans}), passivisation (\textit{the beans were spilled}), and abstract over articles (\textit{spill beans}).

\begin{figure}[h]
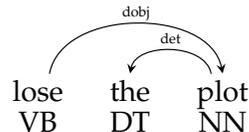

	\centering
	\vspace{5em}
	\begin{dependency}[theme=simple, transform canvas={scale=1.1}]
		\begin{deptext}[column sep=0.4cm]
			lose \& the \& plot \\
			VB \& DT \& NN \\
		\end{deptext}
		\depedge{1}{3}{dobj}
		\depedge{3}{2}{det}
	\end{dependency}
	\vspace{0.5em}
	\caption{Automatic dependency parse of the PIE \textit{lose the plot}.}
	\label{fig:parse-pie-lose-plot}
	\vspace{1em}
\end{figure}

\begin{figure}[h]
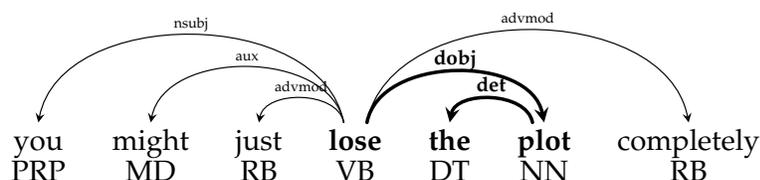

	\centering
	\vspace{5em}
	\begin{dependency}[theme=simple, transform canvas={scale=1.1}]
		\begin{deptext}[column sep=0.4cm]
			you \& might \& just \& \textbf{lose} \& \textbf{the} \& \textbf{plot} \& completely \\
			PRP \& MD \& RB \& VB \& DT \& NN \& RB \\
		\end{deptext}
		\depedge{4}{1}{nsubj}
		\depedge{4}{2}{aux}
		\depedge{4}{3}{advmod}
		\depedge[edge style={very thick}, label style={font=\bfseries}]{4}{6}{dobj}
		\depedge[edge style={very thick}, label style={font=\bfseries}]{6}{5}{det}
		\depedge{4}{7}{advmod}
	\end{dependency}
	\vspace{0.5em}
	\caption{Automatic dependency parse of the sentence `you might just lose the plot completely', which contains the PIE \textit{lose the plot}. From BNC document CH1, sentence 829. Sentence shortened for display convenience.}
	\label{fig:parse-sent-lose-plot}
	\vspace{1em}
\end{figure}

In this method, PIEs are extracted using the assumption that any sentence which contains the lemmata of the words in the PIE, in the same dependency relations as in the PIE, contains an instance of the PIE type in question. More concretely, this means that the parse of the sentence should contain the parse tree of the PIE as a subtree. This is illustrated in Figure~\ref{fig:parse-pie-lose-plot}, which shows the parse tree for the PIE \textit{lose the plot}, parsed without context. Note that this is a subtree of the parse tree for the sentence `you might just lose the plot completely', which is shown in Figure~\ref{fig:parse-sent-lose-plot}. Since the sentence parse contains the parse of the PIE, we can conclude that the sentence contains an instance of that PIE and extract the span of the PIE instance.

All PIEs are parsed in isolation, based on the assumption that all PIEs can be parsed, since they are almost always well-formed phrases. However, not all PIEs will be parsed correctly, especially since there is no context to resolve ambiguity. Errors tend to occur at the part-of-speech level, where, for example, verb-object combinations like \textit{jump ship} and \textit{touch wood} are erroneously tagged as noun-noun compounds. An analysis of the impact of parser error on PIE extraction performance is presented in Section~\ref{sec:extraction-analysis}. Initially, we use the Spacy parser for parsing both the PIEs and the sentences.

Next, the sentence is parsed, and the lemma of the top node of the parsed PIE is matched against the lemmata of the sentence parse. If a match is found, the parse tree of the PIE is matched against the subtree of the matching sentence parse node. If the whole PIE parse tree matches, the span ranging from the first PIE token to the last is extracted. This span can thus include words that are not directly part of the PIE's dictionary form, in order to account for insertions like \textit{ships \textbf{were} jumped} for \textit{jump ship}, or \textit{have a \textbf{big} heart} for \textit{have a heart}.

During the matching, articles (\textit{a}/\textit{an}/\textit{the}) are ignored\footnote{As articles can be inherent parts of idiomatic expressions, we have also tested our method taking articles into account during matching. However, results were lower overall than when ignoring articles. When matching articles, the regular parsing-based method achieves an F1-score of 84.56\%, and the in-context parsing-based method achieves an F1-score of 86.43\%.}, and passivisation is accounted for with a special rule. In addition, a number of special cases are dealt with. These are PIEs containing \textit{someone('s)}, \textit{something('s)}, \textit{one's}, or \textit{\textemdash}. These words are used in PIEs as placeholders for a generic possessor (\textit{someone's/something's/one's}), generic object (\textit{someone/something}), or any word of the right PoS (\textit{\textemdash}). 

For \textit{someone's}, and \textit{something's}, we match any possessive pronoun, or (proper) noun~+ possessive marker. For \textit{one's}, only possessive pronouns are matched, since this is a placeholder for reflexive possessors. For \textit{someone} and \textit{something}, any non-possessive pronoun or (proper) noun is matched.

For \textit{\textemdash} wildcards, any word can be matched, as long as it has the right relation to the right head. An additional challenge with these wildcards is that PIEs containing them cannot be parsed, e.g. \textit{too \textemdash\ for words} is not parseable. This is dealt with by substituting the \textemdash\ by a PoS-ambiguous word, such as \textit{fine}, or \textit{back}. 

\begin{figure}[h]
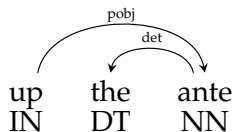

	\centering
	\vspace{5em}
	\begin{dependency}[theme=simple, transform canvas={scale=1.1}]
		\begin{deptext}[column sep=0.4cm]
			up \& the \& ante \\
			IN \& DT \& NN \\
		\end{deptext}
		\depedge{1}{3}{pobj}
		\depedge{3}{2}{det}
	\end{dependency}
	\vspace{0.5em}
	\caption{Automatic dependency parse of the PIE \textit{up the ante}.}
	\label{fig:parse-pie-no-labels}
	\vspace{1em}
\end{figure}

\begin{figure}[h]
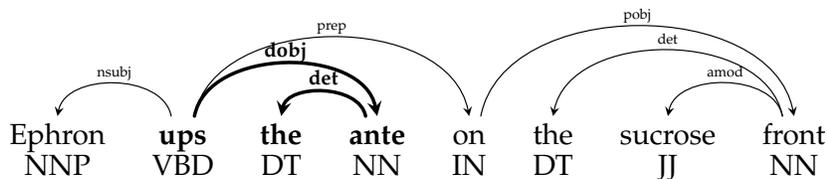

	\centering
	\vspace{5em}
	\begin{dependency}[theme=simple, transform canvas={scale=1.1}]
		\begin{deptext}[column sep=0.4cm]
			Ephron \& \textbf{ups} \& \textbf{the} \& \textbf{ante} \& on \& the \& sucrose \& front \\
			NNP \& VBD \& DT \& NN \& IN \& DT \& JJ \& NN \\
		\end{deptext}
		\depedge{2}{1}{nsubj}
		\depedge[edge style={very thick}, label style={font=\bfseries}]{2}{4}{dobj}
		\depedge[edge style={very thick}, label style={font=\bfseries}]{4}{3}{det}
		\depedge{2}{5}{prep}
		\depedge{5}{8}{pobj}
		\depedge{8}{6}{det}
		\depedge{8}{7}{amod}
	\end{dependency}
	\vspace{0.5em}
	\caption{Automatic dependency parse of the sentence `Ephron ups the ante on the sucrose front', which contains the PIE \textit{up the ante}. From BNC document CBC, sentence 7022. Sentence shortened for display convenience.}
	\label{fig:parse-sent-no-labels}
	\vspace{1em}
\end{figure}

Two optional features are added to the parser-based method with the goal of making it more robust to parser errors: generalising over dependency relation labels, and generalising over dependency relation direction. We expect this to increase recall at the cost of precision. In the first \textit{no labels} setting, we match parts of the parse tree which have the same head lemma and the same dependent lemma, regardless of the relation label. An example of this is Figure~\ref{fig:parse-pie-no-labels}, which has the wrong relation label between \textit{up} and \textit{ante}. If labels are ignored, however, we can still extract the PIE instance in Figure~\ref{fig:parse-sent-no-labels}, which has the correct label. In the \textit{no directionality} setting, relation labels are also ignored, and in addition the directionality of the relation is ignored, that is, we allow for the reversal of heads and dependents. This benefits performance in a case like Figure~\ref{fig:parse-sent-no-direction}, which has \textit{stock} as the head of \textit{laughing} in a \textit{compound} relation, whereas the parse of the PIE (Figure~\ref{fig:parse-pie-no-direction}) has \textit{laughing} as the head of \textit{stock} in a \textit{dobj} relation. 

Note that similar settings were implemented by \citet{Savary2017a}, who detect literal uses of VMWEs using a parser-based method with either full labelled dependencies, unlabelled dependencies, or directionless unlabelled dependencies (which they call BagOfDeps). They find that recall increases when less restrictions on the dependencies are used, but that this does not hurt precision, as we would expect. However, we cannot draw too many conclusions from these results due to the small size of their evaluation set, which consists of just 72 literal VMWEs in total.

\begin{figure}[h]
	\centering
	\vspace{5em}
	\begin{dependency}[theme=simple, transform canvas={scale=1.1}]
		\begin{deptext}[column sep=0.4cm]
			the \& commission \& will \& be \& a \& \textbf{laughing} \& \textbf{stock} \\
			DT \& NN \& MD \& VB \& DT \& VBG \& NN \\
		\end{deptext}
		\depedge{2}{1}{det}
		\depedge{4}{2}{nsubj}
		\depedge{4}{3}{aux}
		\depedge{4}{7}{attr}
		\depedge{7}{5}{det}
		\depedge[edge style={very thick}, label style={font=\bfseries}]{7}{6}{compound}
	\end{dependency}
	\vspace{0.5em}
	\caption{Automatic dependency parse of the sentence `the commission will be a laughing stock', which contains the PIE \textit{laughing stock}. From BNC document A69, sentence 487. Sentence shortened for display convenience.}
	\label{fig:parse-sent-no-direction}
	\vspace{1em}
\end{figure}

\begin{figure}[h]
	\centering
	\vspace{5em}
	\begin{dependency}[theme=simple, transform canvas={scale=1.1}]
		\begin{deptext}[column sep=0.4cm]
			laughing \& stock \\
			VBG \& NN \\
		\end{deptext}
		\depedge{1}{2}{dobj}
	\end{dependency}
	\vspace{0.5em}
	\caption{Automatic dependency parse of the PIE \textit{laughing stock}.}
	\label{fig:parse-pie-no-direction}
	\vspace{1em}
\end{figure}

\begin{figure}[ht]
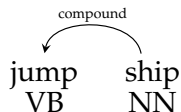

	\centering
	\vspace{5em}
	\begin{dependency}[theme=simple, transform canvas={scale=1.1}]
		\begin{deptext}[column sep=0.4cm]
			jump \& ship \\
			VB \& NN \\
		\end{deptext}
		\depedge{2}{1}{compound}
	\end{dependency}
	\vspace{0.5em}
	\caption{Automatic dependency parse of the PIE \textit{jump ship}.}
	\label{fig:parse-pie-no-context}
	\vspace{1em}
\end{figure}

\begin{figure}[ht]
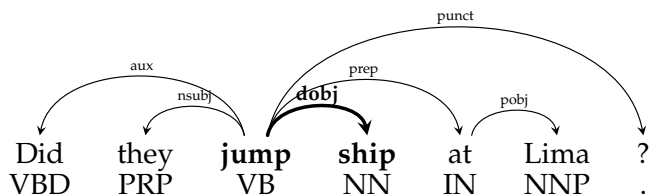

	\centering
	\vspace{5em}
	\begin{dependency}[theme=simple, transform canvas={scale=1.1}]
		\begin{deptext}[column sep=0.4cm]
			Did \& they \& \textbf{jump} \& \textbf{ship} \& at \& Lima \& ? \\
			VBD \& PRP \& VB \& NN \& IN \& NNP \& . \\
		\end{deptext}
		\depedge{3}{1}{aux}
		\depedge{3}{2}{nsubj}
		\depedge[edge style={very thick}, label style={font=\bfseries}]{3}{4}{dobj}
		\depedge{3}{5}{prep}
		\depedge{5}{6}{pobj}
		\depedge{3}{7}{punct}
	\end{dependency}
	\vspace{0.5em}
	\caption{Automatic dependency parse of the extracted sentence `Did they jump ship at Lima?' containing the PIE \textit{jump ship}.}
	\label{fig:parse-pie-context}
	\vspace{1em}
\end{figure}

\paragraph{In-Context Parsing} Since the parser-based method parses PIEs without any context, it often finds an incorrect parse, as for \textit{jump ship} in Figure~\ref{fig:parse-pie-no-context}. As such, we add an option to the method that aims to increase the number of correct parses by parsing the PIE within context, that is, within a sentence. This can greatly help to disambiguate the parse, as in Figure~\ref{fig:parse-pie-context}. If the number of correct parses goes up, the recall of the extraction method should also increase. Naturally, it can also be the case that a PIE is parsed correctly without context, and incorrectly with context. However, we expect the gains to outweigh the losses.

The challenge here is thus to collect example sentences containing the PIE. Since the whole point of this work is to extract PIEs from raw text, this provides a catch-22-like situation: we need to extract a sentence containing a PIE in order to extract sentences containing a PIE. 

The workaround for this problem is to use the exact string matching method with the dictionary form of the PIE and a very large plain text corpus to gather example sentences. By only considering the exact dictionary form we both simplify the finding of example sentences and the extraction of the PIE's parse from the sentence parse.

In case multiple example sentences are found, the shortest sentence is selected, since we assume it is easiest to parse. This is also the reason we make use of very large corpora, to increase the likelihood of finding a short, simple sentence. The example sentence extraction method is modified in such a way that sentences where the PIE is used meta-linguistically in quotes, e.g. ``the well-known English idiom `to spill the beans' has no equivalents in other languages'', are excluded, since they do not provide a natural context for parsing. When no example sentence can be found in the corpus, we back-off to parsing the PIE without context. After a parse has been found for each PIE (i.e. with or without context), the method proceeds identically to the regular parser-based method.

We make use of the combination of two large corpora for the extraction of example sentences: the English Wikipedia\footnote{https://dumps.wikimedia.org/enwiki/20160113/}, and ukWaC \citep{UKWAC}. For the Wikipedia corpus, we use a dump (13-01-2016) of the English-language Wikipedia, and remove all Wikipedia markup. This is done using WikiExtractor\footnote{http://medialab.di.unipi.it/wiki/Wikipedia\_Extractor}. The resulting files still contain some mark-up, which is removed heuristically. The resulting corpus contains mostly clean, raw, untokenized text, numbering approximately 1.78 billion tokens.

As for ukWaC, all XML-markup was removed, and the corpus is converted to a one-sentence-per-line format. UkWaC is tokenized, which makes it difficult for a simple string match method to find PIEs containing punctuation, for example \textit{day in, day out}. Therefore, all spaces before commas, apostrophes, and sentence-final punctuation are removed. The resulting corpus contains approximately 2.05 billion tokens, making for a total of 3.83 billion tokens in the combined ukWaC and Wikipedia corpus. 

\subsection{Results}
\label{sec:extraction-results}	
In order to determine which of the methods described previously produces the highest quality extraction of potentially idiomatic expressions, we evaluate them, in various settings, on the corpus described in Section~\ref{sec:corpus-annotation}.

For parser-based extraction, systems with and without in-context parsing, ignoring labels, and ignoring directionality are tested. For the three string-based extraction methods, varying numbers of intervening words and case sensitivity are evaluated. Evaluation is done using the development set, consisting of 22 documents and 1112 PIE candidates, and the test set, which consists of 23 documents and 1127 PIE candidates. For each method the best set of parameters and/or options is determined using the development set, after which the best variant by F1-score of each method is evaluated on the test set.

Since these documents in the corpus are exhaustively annotated for PIEs (see Section~\ref{sec:evaluation-of-extraction}), we can calculate true and false positives, and false negatives, and thus precision, recall and F1-score. The exact spans are ignored, because the spans annotated in the evaluation corpus are not completely reliable. These were automatically generated during candidate extraction, as described in Section~\ref{sec:candidate-extraction}. Rather, we count an extraction as a true positive if it finds the correct PIE type in the correct sentence.

Note that we judge the system with the highest F1-score to be the best-performing system, since it is a clear and objective criterion. However, when using the system in practice, the best performance depends on the goal. When used as a preprocessing step for PIE disambiguation, the system with the highest F1-score is perhaps the most suitable, but as a corpus building tool, one might want to sacrifice some precision for an increase in recall. This helps to get the most comprehensive annotation of PIEs possible, without overloading the annotators with false extractions (i.e. non-PIEs), by maintaining high precision.

\begin{table}[hbt]	
	\setlength{\tabcolsep}{2.5pt}
	\caption{PIE extraction performance in terms of precision, recall, and F1-score of the three string-based systems (exact, fuzzy, and inflectional), with different options, on the development set. The number of words indicates the number of intervening words allowed between the parts of the PIE for matching to occur. \textit{CS} indicates case-sensitive string matching. The best score for each metric and system is in \textbf{bold}.}
	\label{table:string-dev-results}
	\centering
	\begin{tabular}{lrrr|rrr|rrr|rrr}
		\toprule
		& \multicolumn{3}{c}{0 words} & \multicolumn{3}{c}{1 word} & \multicolumn{3}{c}{2 words} & \multicolumn{3}{c}{3 words} \\
		& P & R & F1 & P & R & F1 & P & R & F1 & P & R & F1 \\
		\midrule
		Exact & 92.80 & 59.19 & 72.28 & 90.73 & 66.54 & \textbf{76.78} & 83.48 & 67.83 & 74.85 & 77.29 & \textbf{68.20} & 72.46 \\
		Exact-CS & \textbf{97.35} & 54.04 & 69.50 & 94.83 & 60.66 & 73.99 & 87.73 & 61.76 & 72.49 & 81.25 & 62.13 & 70.42 \\ \midrule
		Fuzzy & 64.26 & 68.75 & 66.43 & 37.53 & 76.65 & 50.39 & 21.50 & 77.39 & 33.65 & 14.42 & \textbf{77.02} & 24.29 \\
		Fuzzy-CS & \textbf{75.33} & 62.87 & 68.54 & 69.51 & 70.40 & \textbf{69.95} & 59.06 & 71.88 & 64.84 & 51.17 & 72.24 & 59.91 \\ \midrule
		Inflect & 89.79 & 71.14 & 79.38 & 87.10 & 80.70 & \textbf{83.78} & 80.11 & 82.90 & 81.48 & 73.66 & \textbf{83.27} & 78.17 \\
		Inflect-CS & \textbf{93.90} & 65.07 & 76.87 & 90.74 & 73.90 & 81.46 & 83.94 & 75.92 & 79.73 & 77.57 & 76.29 & 76.92 \\
		\bottomrule
	\end{tabular}

\end{table}

\begin{table}[hbt]
	\setlength{\tabcolsep}{4pt}
	\centering	
	\caption{PIE extraction performance in terms of precision, recall, and F1-score of the parser-based system, with different options, on the development set. The best score for each metric is in \textbf{bold}.}
	\label{table:parse-dev-results}
	\begin{tabular}{lrrr|rrr|rrr}
		\toprule
		& \multicolumn{3}{c}{Regular} & \multicolumn{3}{c}{No Labels} & \multicolumn{3}{c}{No Directionality} \\
		& P & R & F1 & P & R & F1 & P & R & F1 \\ \midrule
		Parsing & \textbf{90.83} & 80.15 & 85.16 & 80.00 & 84.56 & 82.22 & 51.40 & 87.68 & 64.81 \\
		In-context-parsing & 89.79 & 87.32 & \textbf{88.54} & 55.29 & 89.34 & 68.31 & 39.61 & \textbf{90.44} & 55.10 \\	
		\bottomrule		
	\end{tabular}
\end{table}

The results for each system on the development set are presented in Tables~\ref{table:string-dev-results} and~\ref{table:parse-dev-results}. Generally, results are in line with expectations: (the best) parse-based methods are better than (the best) string-based methods, and within string-based methods, inflectional matching works best. The same goes for the different settings: case-sensitivity increases precision at the cost of recall, allowing intervening words increases recall at the cost of precision, and the same goes for the \textit{no labels} and \textit{no directionality} options for parser-based extraction. Overall, in-context parser-based extraction works best, with an F1 of 88.54\%, whereas fuzzy matching does very poorly.

Within string-based methods, exact matching has the highest precision, but low recall. Fuzzy matching increases recall at a disproportionately large precision cost, whereas inflectional matching combines the best of both worlds and has high recall at a small loss in precision. For the parser-based system, it is notable that parsing idioms within context yields a clear overall improvement by greatly improving recall at a small cost in precision. 


\begin{table}[hbt]
	\centering	
	\caption{PIE extraction performance in terms of precision, recall, and F1-score of the best variant by F1-score of each of the four systems, on the test set. \textit{CS} indicates case-sensitive string matching. The best score for each metric is in \textbf{bold}. }
	\label{table:test-results}
	\begin{tabular}{lrrr}
		\toprule
		& Precision & Recall & F1-score \\ \midrule
		Exact-1Word & \textbf{92.66} & 59.88 & 72.75 \\
		Fuzzy-CS-1Word & 60.19 & 61.86 & 61.01 \\
		Inflect-1Word & 87.76 & 73.72 & 80.13 \\
		Parsing-InContext & 90.78 & \textbf{87.55} & \textbf{89.13} \\
		\bottomrule
	\end{tabular}
\end{table}

We evaluate the best variant of each system, as determined by F1-score, on the test set. This gives us an indication of whether the system is robust enough, or was overfitted on the development data. Results on the test set are shown in Table~\ref{table:test-results}. On average, the results are lower than the results on the development set. The string-based methods perform clearly worse, with drops of about 4\% F1-score for exact and inflectional match, and a large drop of almost 9\% F1-score for fuzzy matching. The parser-based method, on the other hand, is more robust, with a small 0.59\% increase in F1-score on the test set.

\subsection{Analysis}	
\label{sec:extraction-analysis}

Broadly speaking, the PIE extraction systems presented above perform in line with expectations. It is nevertheless useful to see where the best-performing system misses out, and where improvements like in-context parsing help performance. 

We analyse the shortcomings of the in-context parser-based system by looking at the false positives and false negatives on the development set. We consider the output of the system with best overall performance, since it will provide the clearest picture. 

The system extracts 529 PIEs in total, of which 54 are false extractions (false positives), and it misses 69 annotated PIE instances (false negatives). Most false positives stem from the system's failure to capture nuances of PIE annotation. This includes cases where PIEs contain, or are part of, proper nouns (Example~\ref{example:pie-proper-noun}), PIEs that are part of coordination constructions (Example~\ref{example:pie-coordination}), and incorrect attachments (Example~\ref{example:pie-attachment}). Among these errors, sentences containing proper nouns are an especially frequent problem.

\ex.
\label{example:pie-proper-noun}
Drama series include [..] airline security thrills \textbf{in Cleared} For Takeoff and Head Over Heels [..] (\textit{in the clear} - BNC - document CBC - sentence 5177)

\ex.
\label{example:pie-coordination}
They prefer silk, satin or lace underwear \textbf{in tasteful black} or ivory. (\textit{in the black} - BNC - document CBC - sentence 14673)


\ex.
\label{example:pie-attachment}
[..] `I saw this chap make something \textbf{out of an ordinary piece of wood} — he fashioned it into an exquisite work of art.' (\textit{out of the woods} - BNC - document ABV - sentence 1300)

The main cause of false negatives are errors made by the parser. In order to correctly extract a PIE from a sentence, both the PIE and the sentence have to be parsed correctly, or at least parsed in the same way. This means a missed extraction can be caused by a wrong parse for the PIE or a wrong parse for the sentence. These two error types form the largest class of false negatives. Since some PIE types are rather frequent, a wrong parse for a single PIE type can potentially lead to a large number of missed extractions. 

It is not surprising that the parser makes many mistakes, since idioms often have unusual syntactic constructions (e.g. \textit{come a cropper}) and contain words where default part-of-speech tags lead to the wrong interpretation (e.g. \textit{round} is a preposition in \textit{round the bend}, not a noun or adjective). This is especially true when idioms are parsed without context, and hence, where in-context parsing provides the largest benefit: the number of PIEs which are parsed incorrectly drops, which leads to F1-scores on those types going from 0\% to almost 100\% (e.g. \textit{in light of} and \textit{ring a bell}). Since parser errors are the main contributor to false negatives, hurting recall, we can observe that parsing idioms in context serves to benefit only recall, by 7 percentage points, at only a small loss in precision.

\begin{figure}[ht]
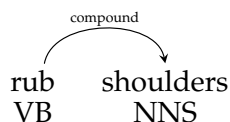

	\centering
	\vspace{5em}
	\begin{dependency}[theme=simple, transform canvas={scale=1.1}]
		\begin{deptext}[column sep=0.4cm]
			rub \& shoulders \\
			VB \& NNS \\
		\end{deptext}
		\depedge{1}{2}{compound}
	\end{dependency}
	\vspace{0.5em}
	\caption{Automatic dependency parse of the PIE \textit{rub shoulders}.}
	\label{fig:rub-shoulders}
	\vspace{1em}
\end{figure}

We find that adding context mainly helps for parsing expressions which are structurally relatively simple, but still ambiguous, such as \textit{rub shoulders}, \textit{laughing stock}, and \textit{round the bend}. Compare, for example, the parse trees for \textit{laughing stock} in isolation and within the extracted context sentence in Figures~\ref{fig:rub-shoulders} and~\ref{fig:rub-shoulders-context}. When parsed in isolation, the relation between the two words is incorrectly labelled as a compound relation, whereas in context it is correctly labelled as a direct object relation. Note however, that for the most difficult PIEs, embedding them in a context does solve the parsing problem: a syntactically odd phrase is hard to phrase (e.g. \textit{for the time being}), and a syntactically odd phrase in a sentence makes for a syntactically odd sentence that is still hard to parse (e.g. `London for the time being had been abandoned.'). Finding example sentences turned out not to be a problem, since appropriate sentences were found for 559 of 591 PIE types. 

\begin{figure}[ht]
	\centering
	\vspace{5em}
	\begin{dependency}[theme=simple, transform canvas={scale=1.1}]
		\begin{deptext}[column sep=0.5cm]
			Each \& day \& they \& \textbf{rub} \& \textbf{shoulders} \& with \& death \& . \\
			DT \& NN \& PRP \& VBP \& NNS \& IN \& NN \& . \\
		\end{deptext}
		\depedge{2}{1}{det}
		\depedge{4}{2}{npadvmod}
		\depedge{4}{3}{nsubj}
		\depedge[edge style={very thick}, label style={font=\bfseries}]{4}{5}{dobj}
		\depedge{4}{6}{prep}
		\depedge{6}{7}{pobj}
		\depedge{4}{8}{punct}
	\end{dependency}
	\vspace{0.5em}
	\caption{Automatic dependency parse of the extracted sentence `Each day they rub shoulders with death.' containing the PIE \textit{rub shoulders}.}
	\label{fig:rub-shoulders-context}
	\vspace{1em}
\end{figure}


An alternative method for reducing parser error is to use a different, better parser. The Spacy parser was mainly chosen for implementation convenience and speed, and there are parsers which have better performance, as measured on established parsing benchmarks. To investigate the effectiveness of this method, we used the Stanford Neural Dependency Parser \citep{Chen2014} to extract PIEs in the regular parsing, in-context parsing and the \textit{no labels} settings. In all cases, using the Stanford parser yielded worse extraction performance than the Spacy parser. A possible explanation for why a supposedly better parser performs worse here is that parsers are optimised and trained to do well on established benchmarks, which consist of complete sentences, often from news texts. This does not necessarily correlate with parsing performance on short (sentences containing) idiomatic phrases. As such, we cannot assume that better overall parsing performance implies PIE extraction performance.

It should be noted that, when assessing the quality of PIE extraction performance, the parser-based methods are sensitive to specific PIE types. That is, if a single PIE type is parsed incorrectly, then it is highly probable that all instances of that type are missed. If this type is also highly frequent, this means that a small change in actual performance yields a large change in evaluation scores. Our goal is to have a PIE extraction system that is robust across all PIE types, and thus the current evaluation setting does not align exactly with our aim.

Splitting out performance per PIE type reveals whether there is indeed a large variance in performance across types. Table~\ref{table:pie-type-performance} shows the 25 most frequent PIE types in the corpus, and the performance of the in-context-parsing-based system on each. 
Except two cases (\textit{in the black} and \textit{round the bend}), we see that the performance is in the 80--100\% range, even showing perfect performance on the majority of types. 
%


\begin{table}[hbt]
	\centering
	\caption{Extraction performance of the in-context-parsing-based system on each of the 25 most frequent PIE types in the corpus.}
	\label{table:pie-type-performance}
	\begin{tabular}{lrrrr}
		\toprule
		PIE Type & Count & Precision & Recall & F1-score \\ \midrule
\textit{on the ground} & 48 & 96.00 & 100.00 & 97.96 \\
\textit{on board} & 24 & 100.00  & 83.33  & 90.91 \\
\textit{on the cards}  & 18  & 94.74  & 100.00  & 97.30 \\
\textit{at sea}  & 15  & 93.33  & 93.33  & 93.33 \\
\textit{in someone's pocket}  & 13  & 90.91  & 76.92  & 83.33 \\
\textit{in the hole}  & 9  & 100.00  & 100.00  & 100.00 \\
\textit{all along}  & 9  & 100.00  & 100.00  & 100.00 \\
\textit{all over the place}  & 9  & 100.00  & 100.00  & 100.00 \\
\textit{under fire}  & 8  & 100.00  & 87.50  & 93.33 \\
\textit{in light of}  & 8  & 100.00  & 87.50  & 93.33 \\
\textit{on the level}  & 8  & 100.00  & 100.00  & 100.00 \\
\textit{over the top} & 7  & 100.00  & 100.00  & 100.00 \\
\textit{on edge}  & 7  & 100.00  & 100.00  & 100.00 \\
\textit{at the end of the day}  & 7  & 100.00  & 100.00  & 100.00 \\
\textit{ring a bell}  & 6  & 100.00  & 66.67  & 80.00 \\
\textit{in the bag}  & 6  & 85.71  & 100.00  & 92.31 \\
\textit{in the running } & 6  & 100.00  & 83.33  & 90.91 \\
\textit{up for grabs}  & 6  & 100.00  & 100.00  & 100.00 \\
\textit{on the rocks } & 5  & 100.00  & 100.00  & 100.00 \\
\textit{in the black } & 5  & 40.00  & 40.00  & 40.00 \\
\textit{out of the blue } & 5  & 100.00  & 100.00  & 100.00 \\
\textit{round the bend } & 5  & 100.00  & 40.00  & 57.14 \\
\textit{behind bars} & 5  & 100.00  & 100.00  & 100.00 \\
\textit{have a go}  & 5  & 71.43  & 100.00  & 83.33 \\
\textit{turn the corner}  & 4  & 100.00  & 100.00  & 100.00 \\
		\bottomrule	
	\end{tabular}
\end{table}

For none of the types do we see low precision paired with high recall, which indicates that the parser never matches a highly frequent non-PIE phrase. For the system with the \textit{no labels} and \textit{no-directionality} options (per-type numbers not shown here), however, this does occur. For example, ignoring the labels for the parse of the PIE \textit{have a go} leads to the erroneous matching of many sentences containing a form of \textit{have to go}, which is highly frequent, thus leading to a large drop in precision.  

Although performance is stable across the most frequent types, among the less frequent types it is more spotty. This hurts overall performance, and there are potential gains in mitigating the poor performance on these types, such as \textit{for the time being}. At the same time, the string matching methods show much more stable performance across types, and some of them do so with very high precision. As such, a combination of two such methods could boost performance significantly. If we use a high-precision string match-based method, such as the exact string match variant with a precision of 97.35\%, recall could be improved for the wrongly parsed PIE types, without a significant loss of precision.

We experiment with two such combinations, by simply taking the union of the sets of extracted idioms of both systems, and filtering out duplicates. Results are shown in Table~\ref{table:combined-dev-results}. Both combinations show the expected effect: a clear gain in recall at a minimal loss in precision. Compared to the in-context-parsing-based system, the combination with exact string matching yields a gain in recall of over 6\%, and the combination with inflectional string matching yields an even bigger gain of almost 8\%, at precision losses of 0.6\% and 0.8\%, respectively. This indicates that the systems are very much complementary in the PIEs they extract. It also means that, when used in practice, combining inflectional string matching and parse-based extraction is the most reliable configuration.

\begin{table}[hbt]
	\centering	
	\caption{PIE extraction performance of the combined output (union) of a string-based and a parser-based system, on the development set. \textit{CS} indicates case-sensitive string matching. The best score for each metric is in \textbf{bold}.}
	\label{table:combined-dev-results}
	\begin{tabular}{lrrr}
		\toprule
		& Precision	& Recall & F1-score \\ \midrule
		Parsing-InContext $\cup$ Exact-CS-0Words & \textbf{89.18} & 93.93 & 91.50 \\
		Parsing-InContext $\cup$ Inflect-CS-0Words & 89.00 & \textbf{95.22} & \textbf{92.01} \\
		\bottomrule		
	\end{tabular}
\end{table}

\section{Conclusions and Outlook}
We present an in-depth study on the automatic extraction of potentially idiomatic expressions based on dictionaries. The purpose of automatic dictionary-based extraction is, on the one hand, to function as a pre-extraction step in the building of a large idiom-annotated corpus. On the other hand, it can function as part of an idiom extraction system when combined with a disambiguation component. In both cases, the ultimate goal is to improve the processing of idiomatic expressions within NLP. This work consists of three parts: a comparative evaluation of the coverage of idiom dictionaries, the annotation of a PIE corpus, and the development and evaluation of several dictionary-based PIE extraction methods.

In the first part, we present a study of idiom dictionary coverage, which serves to answer the question of whether a single idiom dictionary, or a combination of dictionaries, can provide good coverage of the set of all English idioms. Based on the comparison of dictionaries to each other, we estimate that the overlap between them is limited, varying from 20\% to 55\%, which indicates a large divergence between the dictionaries. This can be explained by the fact that idioms vary widely by register, genre, language variety, and time period. In our case, it is also likely that the divergence is caused partly by the gap between crowdsourced dictionaries on the one hand, and a dictionary compiled by professional lexicographers on the other. Given these factors, we can conclude that a single dictionary cannot provide even close to complete coverage of English idioms, but that by combining dictionaries from various sources, significant gains can be made. Since `English idioms' are a diffuse and constantly changing set, we have no gold standard to compare to. As such, we conclude that multiple dictionaries should be used when possible, but that we cannot say any anything definitive on the coverage of dictionaries with regard to the complete set of English idioms (which can only be approximated in the first place). A more comprehensive of idiom resources could be made in the future by using more advanced automatic methods for matching, for example by using \citeauthor{Pasquer2018b}'s (\citeyear{Pasquer2018b}) method for measuring expression variability. This would make it easier to evaluate a larger number of dictionaries, since no manual effort would be required.

In the second part, we experiment with the exhaustive annotation of PIEs in a corpus of documents from the BNC.\footnote{The corpus annotations are available at https://github.com/hslh/pie-annotation under a CC-BY-4.0 licence. The source code of the PIE extraction systems is available at https://github.com/hslh/pie-detection, also under a CC-BY-4.0 licence.} Using a set of 591 PIE types, much larger and more varied than in existing resources, we show that it is very much possible to establish a working definition of PIE that allows for a large amount of variation, while still being useful for reliable annotation. This resulted in high inter-annotator agreement, ranging from 0.74 to 0.91 Fleiss' Kappa. This means that we can build a resource to evaluate a wide-range idiom extraction system with relatively little effort. The final corpus of PIEs with sense annotations is publicly available consists of 2,239 PIE candidates, of which 1,050 actual PIEs instances, and contains 278 different PIE types.

Finally, several methods for the automatic extraction of PIE instances were developed and evaluated on the annotated PIE corpus. We tested methods of differing complexity, from simple string match to dependency parse-based extraction. Comparison of these methods revealed that the more computationally complex method, parser-based extraction, works best. Parser-based extraction is especially effective in capturing a larger amount of variation, but is less precise than string-based methods, mostly because of parser error. The best overall setting of this method, which parses idioms within context, yielded an F1-score of 89.13\% on the test set. Parser error can be partly compensated by combining the parse-based method and the inflectional string match method, which yields an F1-score of 92.01\% (on the development set). This aligns well with the findings by \citet{Baldwin2005}, who found that combining simpler and more complex methods improves over just using a simple method case for extracting verb-particle constructions. This level of performance means that we can use the tool in corpus building. This greatly reduces the amount of manual extraction effort involved, while still maintaining a high level of recall. We make the source code for the different systems publicly available.

Note that, although used here in the context of PIE extraction, our methods are equally applicable to other phrase extraction tasks, for example the extraction of light-verb constructions, metaphoric constructions, collocations, or any other type of multiword expression (cf. \citealp{Baldwin2005,Inurrieta2016,Savary2017a}). Similarly, our method can be conceived as a blueprint and extended to languages other than English. For this to be possible,  for any given new language one would need a list of target expressions and, in the case of the parser-based method, a reliable syntactic parser. If this is not the case, the inflectional matching method can be used, which requires only a morphological analyser and generator. Obviously, for languages that are morphologically richer than English, one would need to develop strategies aimed at controlling non-exact matches, so as to enhance recall without sacrificing precision. Previous work on Italian, for example, has shown the feasibility of achieving such balance through controlled pattern matching \cite{nissim2013modeling}. Languages that are typologically very different from English would obviously require a dedicated approach for the matching of PIEs in corpora, but the overall principles of extraction, using language-specific tools, could stay the same.  

Currently, no corpora containing annotation of PIEs exist for languages other than English. However, the PARSEME corpus \cite{Savary2018} already contains idioms (only idiomatic readings) for many languages and would only need annotation of literal usages of idioms to make up a set of PIEs. Paired with the Universal Dependencies project \cite{UD}, which increasingly provides annotated data as well as processing tools for an ever growing number of languages, this seems an excellent starting point for creating PIE resources in multiple languages. 


\starttwocolumn
\bibliography{RandomInteresting,NonLiterals}

\begin{thebibliography}{42}
\expandafter\ifx\csname natexlab\endcsname\relax\def\natexlab#1{#1}\fi

\bibitem[{Ayto(2009)}]{Ayto2009}
Ayto, John, editor. 2009.
\newblock \emph{From the horse's mouth: {Oxford} dictionary of {English}
  Idioms}, 3rd edition.
\newblock Oxford University Press, Oxford; New York.

\bibitem[{Baldwin(2005)}]{Baldwin2005}
Baldwin, Timothy. 2005.
\newblock The deep lexical acquisition of {E}nglish verb-particle
  constructions.
\newblock \emph{Computer Speech and Language}, 19(4):398--414.

\bibitem[{BNC()}]{BNC-XML}
BNC. 2007.
\newblock The {B}ritish {N}ational {C}orpus, version 3 ({BNC} {XML} {E}dition).
\newblock Distributed by Bodleian Libraries, University of Oxford, on behalf of
  the BNC Consortium.

\bibitem[{Boukobza and Rappoport(2009)}]{Boukobza2009}
Boukobza, Ram and Ari Rappoport. 2009.
\newblock Multi-word expression identification using sentence surface features.
\newblock In \emph{Proceedings of the 2009 Conference on Empirical Methods in
  Natural Language Processing}, pages 468--477, Association for Computational
  Linguistics, Singapore.

\bibitem[{Burnard(2007)}]{BNCRef}
Burnard, Lou. 2007.
\newblock Reference guide for the {B}ritish {N}ational {C}orpus ({XML}
  edition).

\bibitem[{Chen and Manning(2014)}]{Chen2014}
Chen, Danqi and Christopher~D. Manning. 2014.
\newblock A fast and accurate dependency parser using neural networks.
\newblock In \emph{Proceedings of the 2014 Conference on Empirical Methods in
  Natural Language Processing ({EMNLP})}, pages 740--750.

\bibitem[{Constant et~al.(2017)Constant, Eryi{\u{g}}it, Monti, Plas, Ramisch,
  Rosner, and Todirascu}]{Constant2017}
Constant, Mathieu, G{\"u}l{\c{s}}en Eryi{\u{g}}it, Johanna Monti, Lonneke
  van~der Plas, Carlos Ramisch, Michael Rosner, and Amalia Todirascu. 2017.
\newblock Survey: Multiword expression processing: A survey.
\newblock \emph{Computational Linguistics}, 43(4):837--892.

\bibitem[{Cook, Fazly, and Stevenson(2008)}]{Cook2008}
Cook, Paul, Afsaneh Fazly, and Suzanne Stevenson. 2008.
\newblock The {VNC-T}okens dataset.
\newblock In \emph{Proceedings of the {LREC} workshop towards a shared task for
  Multiword Expressions}, pages 19--22.

\bibitem[{Fazly, Cook, and Stevenson(2009)}]{Fazly2009}
Fazly, Afsaneh, Paul Cook, and Suzanne Stevenson. 2009.
\newblock Unsupervised type and token identification of idiomatic expressions.
\newblock \emph{Computational Linguistics}, 35(1):61--103.

\bibitem[{Ferraresi et~al.(2008)Ferraresi, Zanchetta, Baroni, and
  Bernardini}]{UKWAC}
Ferraresi, Adriano, Eros Zanchetta, Marco Baroni, and Silvia Bernardini. 2008.
\newblock Introducing and evaluating {ukWaC}, a very large web-derived corpus
  of {E}nglish.
\newblock In \emph{Proceedings of {LREC}}, pages 47--54.

\bibitem[{Finlayson and Kulkarni(2011)}]{Finlayson2011}
Finlayson, Mark and Nidhi Kulkarni. 2011.
\newblock Detecting multi-word expressions improves word sense disambiguation.
\newblock In \emph{Proceedings of the Workshop on Multiword Expressions: from
  Parsing and Generation to the Real World}, pages 20--24, Association for
  Computational Linguistics, Portland, Oregon, USA.

\bibitem[{Fischer and Keil(1996)}]{Fischer1996}
Fischer, Ingrid and Martina Keil. 1996.
\newblock Parsing decomposable idioms.
\newblock In \emph{Proceedings of {COLING}}, pages 388--393.

\bibitem[{Geeraert, Baayen, and Newman(2017)}]{Geeraert2017}
Geeraert, Kristina, R.~Harald Baayen, and John Newman. 2017.
\newblock Understanding idiomatic variation.
\newblock In \emph{Proceedings of the 13th Workshop on Multiword Expressions
  ({MWE} 2017)}, pages 80--90.

\bibitem[{Geoffrey~Nunberg(1994)}]{Nunberg1994}
Geoffrey~Nunberg, Thomas~Wasow, Ivan A.~Sag. 1994.
\newblock Idioms.
\newblock \emph{Language}, 70(3):491--538.

\bibitem[{Gharbieh, Bhavsar, and Cook(2016)}]{Gharbieh2016}
Gharbieh, Waseem, Virendra~C. Bhavsar, and Paul Cook. 2016.
\newblock A word embedding approach to identifying verb-noun idiomatic
  combinations.
\newblock In \emph{Proceedings of the 12th Workshop on {Multiword}
  {Expressions}}, pages 112--118.

\bibitem[{Gong, Bhat, and Viswanath(2016)}]{Gong2016}
Gong, Hongyu, Suma Bhat, and Pramod Viswanath. 2016.
\newblock Geometry of compositionality.
\newblock \emph{CoRR}, abs/1611.09799.

\bibitem[{Graff and Cieri(2003)}]{Gigaword2003}
Graff, David and Christopher Cieri. 2003.
\newblock {English} {Gigaword}.

\bibitem[{Gr\'{e}goire(2009)}]{Gregoire2009}
Gr\'{e}goire, Nicole. 2009.
\newblock \emph{Untangling Multiword Expressions: A study on the representation
  and variation of {D}utch multiword expressions}.
\newblock Ph.D. thesis, Universiteit Utrecht.

\bibitem[{Gulland and Hinds-Howell(2002)}]{Gulland2002}
Gulland, Daphne~M. and David Hinds-Howell, editors. 2002.
\newblock \emph{The {Penguin} dictionary of {E}nglish Idioms}, 2nd edition.
\newblock Penguin Books, London.

\bibitem[{Ide and V\'{e}ronis(1994)}]{Ide1994}
Ide, Nancy and Jean V\'{e}ronis. 1994.
\newblock Machine readable dictionaries: What have we learned, where do we go?
\newblock In \emph{Proceedings of the International Workshop on the Future of
  Lexical Research}, pages 137--146, Beijing, China.

\bibitem[{I{\~n}urrieta et~al.(2016)I{\~n}urrieta, de~Ilarraza, Labaka,
  Sarasola, Aduriz, and Carroll}]{Inurrieta2016}
I{\~n}urrieta, Uxoa, Arantza~D\'{i}az de~Ilarraza, Gorka Labaka, Kepa Sarasola,
  Itziar Aduriz, and John Carroll. 2016.
\newblock Using linguistic data for {E}nglish and {S}panish verb-noun
  combination identification.
\newblock In \emph{Proceedings of {COLING}}, pages 857--867.

\bibitem[{Isabelle, Cherry, and Foster(2017)}]{Isabelle2017}
Isabelle, Pierre, Colin Cherry, and George Foster. 2017.
\newblock A challenge set approach to evaluating machine translation.
\newblock In \emph{{Proceedings of the 2017 Conference on Empirical Methods in
  Natural Language Processing ({EMNLP})}}, pages 2476--2486, Association for
  Computational Linguistics.

\bibitem[{Korkontzelos et~al.(2013)Korkontzelos, Zesch, Zanzotto, and
  Biemann}]{Korkontzelos2013}
Korkontzelos, Ioannis, Torsten Zesch, Fabio~Massimo Zanzotto, and Chris
  Biemann. 2013.
\newblock {SemEval-2013} task 5: Evaluating phrasal semantics.
\newblock In \emph{Proceedings of {SemEval}}, pages 39--47.

\bibitem[{Minnen, Carroll, and Pearce(2001)}]{Minnen2001}
Minnen, Guido, John Carroll, and Darren Pearce. 2001.
\newblock Applied morphological processing of {English}.
\newblock \emph{Natural Language Engineering}, 7(3):207--223.

\bibitem[{Minugh(2007)}]{Minugh2007}
Minugh, David. 2007.
\newblock The filling in the sandwich: internal modification of idioms.
\newblock In Roberta Facchinetti, editor, \emph{Corpus Linguistics 25 Years
  on}. Rodopi, Amsterdam, pages 205--224.

\bibitem[{Muzny and Zettlemoyer(2013)}]{Muzny2013}
Muzny, Grace and Luke Zettlemoyer. 2013.
\newblock Automatic idiom identification in {Wiktionary}.
\newblock In \emph{Proceedings of {EMNLP}}, pages 1417--1421.

\bibitem[{Nissim and Zaninello(2013)}]{nissim2013modeling}
Nissim, Malvina and Andrea Zaninello. 2013.
\newblock Modeling the internal variability of multiword expressions through a
  pattern-based method.
\newblock \emph{ACM Transactions on Speech and Language Processing (TSLP)},
  10(2):7.

\bibitem[{Nivre et~al.(2017)Nivre, Agi{\'c}, Ahrenberg, Aranzabe, Asahara,
  Atutxa, Ballesteros, Bauer, Bengoetxea, Bhat, Bick, Bosco, Bouma, Bowman,
  Candito, Cebiro{\u g}lu~Eryi{\u g}it, Celano, Chalub, Choi, {\c
  C}{\"o}ltekin, Connor, Davidson, de~Marneffe, de~Paiva, Diaz~de Ilarraza,
  Dobrovoljc, Dozat, Droganova, Dwivedi, Eli, Erjavec, Farkas, Foster, Freitas,
  Gajdo{\v s}ov{\'a}, Galbraith, Garcia, Ginter, Goenaga, Gojenola,
  G{\"o}k{\i}rmak, Goldberg, G{\'o}mez~Guinovart, Gonz{\'a}les~Saavedra,
  Grioni, Gr{\= u}z{\={\i}}tis, Guillaume, Habash, Haji{\v c}, H{\`a}~M{\~y},
  Haug, Hladk{\'a}, Hohle, Ion, Irimia, Johannsen, J{\o}rgensen, Ka{\c
  s}{\i}kara, Kanayama, Kanerva, Kotsyba, Krek, Laippala, L{\^e}~H{\`{\^o}}ng,
  Lenci, Ljube{\v s}i{\'c}, Lyashevskaya, Lynn, Makazhanov, Manning, M{\u
  a}r{\u a}nduc, Mare{\v c}ek, Mart{\'{\i}}nez~Alonso, Martins, Ma{\v s}ek,
  Matsumoto, {McDonald}, Missil{\"a}, Mititelu, Miyao, Montemagni, More, Mori,
  Moskalevskyi, Muischnek, Mustafina, M{\"u}{\"u}risep, Nguy{\~{\^e}}n~Th{\d
  i}, Nguy{\~{\^e}}n Th{\d i}~Minh, Nikolaev, Nurmi, Ojala, Osenova,
  {\O}vrelid, Pascual, Passarotti, Perez, Perrier, Petrov, Piitulainen, Plank,
  Popel, Pretkalni{\c n}a, Prokopidis, Puolakainen, Pyysalo, Rademaker,
  Ramasamy, Real, Rituma, Rosa, Saleh, Sanguinetti, Saul{\={\i}}te, Schuster,
  Seddah, Seeker, Seraji, Shakurova, Shen, Sichinava, Silveira, Simi,
  Simionescu, Simk{\'o}, {\v S}imkov{\'a}, Simov, Smith, Suhr, Sulubacak,
  Sz{\'a}nt{\'o}, Taji, Tanaka, Tsarfaty, Tyers, Uematsu, Uria, van Noord,
  Varga, Vincze, Washington, {\v Z}abokrtsk{\'y}, Zeldes, Zeman, and Zhu}]{UD}
Nivre, Joakim, {\v Z}eljko Agi{\'c}, Lars Ahrenberg, Maria~Jesus Aranzabe,
  Masayuki Asahara, Aitziber Atutxa, Miguel Ballesteros, John Bauer, Kepa
  Bengoetxea, Riyaz~Ahmad Bhat, Eckhard Bick, Cristina Bosco, Gosse Bouma, Sam
  Bowman, Marie Candito, G{\"u}l{\c s}en Cebiro{\u g}lu~Eryi{\u g}it, Giuseppe
  G.~A. Celano, Fabricio Chalub, Jinho Choi, {\c C}a{\u g}r{\i} {\c
  C}{\"o}ltekin, Miriam Connor, Elizabeth Davidson, Marie-Catherine
  de~Marneffe, Valeria de~Paiva, Arantza Diaz~de Ilarraza, Kaja Dobrovoljc,
  Timothy Dozat, Kira Droganova, Puneet Dwivedi, Marhaba Eli, Toma{\v z}
  Erjavec, Rich{\'a}rd Farkas, Jennifer Foster, Cl{\'a}udia Freitas,
  Katar{\'{\i}}na Gajdo{\v s}ov{\'a}, Daniel Galbraith, Marcos Garcia, Filip
  Ginter, Iakes Goenaga, Koldo Gojenola, Memduh G{\"o}k{\i}rmak, Yoav Goldberg,
  Xavier G{\'o}mez~Guinovart, Berta Gonz{\'a}les~Saavedra, Matias Grioni,
  Normunds Gr{\= u}z{\={\i}}tis, Bruno Guillaume, Nizar Habash, Jan Haji{\v c},
  Linh H{\`a}~M{\~y}, Dag Haug, Barbora Hladk{\'a}, Petter Hohle, Radu Ion,
  Elena Irimia, Anders Johannsen, Fredrik J{\o}rgensen, H{\"u}ner Ka{\c
  s}{\i}kara, Hiroshi Kanayama, Jenna Kanerva, Natalia Kotsyba, Simon Krek,
  Veronika Laippala, Phuong L{\^e}~H{\`{\^o}}ng, Alessandro Lenci, Nikola
  Ljube{\v s}i{\'c}, Olga Lyashevskaya, Teresa Lynn, Aibek Makazhanov,
  Christopher Manning, C{\u a}t{\u a}lina M{\u a}r{\u a}nduc, David Mare{\v
  c}ek, H{\'e}ctor Mart{\'{\i}}nez~Alonso, Andr{\'e} Martins, Jan Ma{\v s}ek,
  Yuji Matsumoto, Ryan {McDonald}, Anna Missil{\"a}, Verginica Mititelu, Yusuke
  Miyao, Simonetta Montemagni, Amir More, Shunsuke Mori, Bohdan Moskalevskyi,
  Kadri Muischnek, Nina Mustafina, Kaili M{\"u}{\"u}risep, Luong
  Nguy{\~{\^e}}n~Th{\d i}, Huy{\`{\^e}}n Nguy{\~{\^e}}n Th{\d i}~Minh, Vitaly
  Nikolaev, Hanna Nurmi, Stina Ojala, Petya Osenova, Lilja {\O}vrelid, Elena
  Pascual, Marco Passarotti, Cenel-Augusto Perez, Guy Perrier, Slav Petrov,
  Jussi Piitulainen, Barbara Plank, Martin Popel, Lauma Pretkalni{\c n}a,
  Prokopis Prokopidis, Tiina Puolakainen, Sampo Pyysalo, Alexandre Rademaker,
  Loganathan Ramasamy, Livy Real, Laura Rituma, Rudolf Rosa, Shadi Saleh,
  Manuela Sanguinetti, Baiba Saul{\={\i}}te, Sebastian Schuster, Djam{\'e}
  Seddah, Wolfgang Seeker, Mojgan Seraji, Lena Shakurova, Mo~Shen, Dmitry
  Sichinava, Natalia Silveira, Maria Simi, Radu Simionescu, Katalin Simk{\'o},
  M{\'a}ria {\v S}imkov{\'a}, Kiril Simov, Aaron Smith, Alane Suhr, Umut
  Sulubacak, Zsolt Sz{\'a}nt{\'o}, Dima Taji, Takaaki Tanaka, Reut Tsarfaty,
  Francis Tyers, Sumire Uematsu, Larraitz Uria, Gertjan van Noord, Viktor
  Varga, Veronika Vincze, Jonathan~North Washington, Zden{\v e}k {\v
  Z}abokrtsk{\'y}, Amir Zeldes, Daniel Zeman, and Hanzhi Zhu. 2017.
\newblock Universal dependencies 2.0.
\newblock {LINDAT}/{CLARIN} digital library at the Institute of Formal and
  Applied Linguistics ({{\'U}FAL}), Faculty of Mathematics and Physics, Charles
  University.

\bibitem[{Pasquer et~al.(2018{\natexlab{a}})Pasquer, Savary, Antoine, and
  Ramisch}]{Pasquer2018b}
Pasquer, Caroline, Agata Savary, Jean-Yves Antoine, and Carlos Ramisch.
  2018{\natexlab{a}}.
\newblock Towards a variability measure for multiword expressions.
\newblock In \emph{Proceedings of the 2018 Conference of the North American
  Chapter of the Association for Computational Linguistics: Human Language
  Technologies, Volume 2 (Short Papers)}, pages 426--432, Association for
  Computational Linguistics, New Orleans, Louisiana.

\bibitem[{Pasquer et~al.(2018{\natexlab{b}})Pasquer, Savary, Ramisch, and
  Antoine}]{Pasquer2018}
Pasquer, Caroline, Agata Savary, Carlos Ramisch, and Jean-Yves Antoine.
  2018{\natexlab{b}}.
\newblock If you{'}ve seen some, you{'}ve seen them all: Identifying variants
  of multiword expressions.
\newblock In \emph{Proceedings of the 27th International Conference on
  Computational Linguistics}, pages 2582--2594, Association for Computational
  Linguistics, Santa Fe, New Mexico, USA.

\bibitem[{Ramisch et~al.(2018)Ramisch, Cordeiro, Savary, Vincze,
  Barbu~Mititelu, Bhatia, Buljan, Candito, Gantar, Giouli, G{\"u}ng{\"o}r,
  Hawwari, I{\~n}urrieta, Kovalevskait{\.e}, Krek, Lichte, Liebeskind, Monti,
  Parra~Escart{\'\i}n, QasemiZadeh, Ramisch, Schneider, Stoyanova, Vaidya, and
  Walsh}]{Ramisch2018}
Ramisch, Carlos, Silvio~Ricardo Cordeiro, Agata Savary, Veronika Vincze,
  Verginica Barbu~Mititelu, Archna Bhatia, Maja Buljan, Marie Candito, Polona
  Gantar, Voula Giouli, Tunga G{\"u}ng{\"o}r, Abdelati Hawwari, Uxoa
  I{\~n}urrieta, Jolanta Kovalevskait{\.e}, Simon Krek, Timm Lichte, Chaya
  Liebeskind, Johanna Monti, Carla Parra~Escart{\'\i}n, Behrang QasemiZadeh,
  Renata Ramisch, Nathan Schneider, Ivelina Stoyanova, Ashwini Vaidya, and
  Abigail Walsh. 2018.
\newblock Edition 1.1 of the parseme shared task on automatic identification of
  verbal multiword expressions.
\newblock In \emph{Proceedings of the Joint Workshop on Linguistic Annotation,
  Multiword Expressions and Constructions (LAW-MWE-CxG-2018)}, pages 222--240,
  Association for Computational Linguistics, Santa Fe, New Mexico, USA.

\bibitem[{Sag et~al.(2002)Sag, Baldwin, Bond, Copestake, and
  Flickinger}]{Sag2002}
Sag, Ivan~A., Timothy Baldwin, Francis Bond, Ann Copestake, and Dan Flickinger.
  2002.
\newblock Multiword expressions: A pain in the neck for {NLP}.
\newblock In \emph{Proceedings of {CICLING}}, pages 1--15.

\bibitem[{Salton, Ross, and Kelleher(2014)}]{Salton2014}
Salton, Giancarlo~D., Robert~J. Ross, and John~D. Kelleher. 2014.
\newblock An empirical study of the impact of idioms on phrase based
  statistical machine translation of {E}nglish to {B}razilian-{P}ortuguese.
\newblock In \emph{Proceedings of the 3rd Workshop on Hybrid Approaches to
  Translation ({HyTra})}, pages 36--41.

\bibitem[{Savary et~al.(2018)Savary, Candito, Mititelu, Bejček, Cap, Čéplö,
  Cordeiro, Eryiğit, Giouli, van Gompel, HaCohen-Kerner, Kovalevskaitė, Krek,
  Liebeskind, Monti, Escartín, van~der Plas, QasemiZadeh, Ramisch, Sangati,
  Stoyanova, and Vincze}]{Savary2018}
Savary, Agata, Marie Candito, Verginica~Barbu Mititelu, Eduard Bejček,
  Fabienne Cap, Slavomír Čéplö, Silvio~Ricardo Cordeiro, Gülşen Eryiğit,
  Voula Giouli, Maarten van Gompel, Yaakov HaCohen-Kerner, Jolanta
  Kovalevskaitė, Simon Krek, Chaya Liebeskind, Johanna Monti, Carla~Parra
  Escartín, Lonneke van~der Plas, Behrang QasemiZadeh, Carlos Ramisch,
  Federico Sangati, Ivelina Stoyanova, and Veronika Vincze. 2018.
\newblock \emph{{PARSEME multilingual corpus of verbal multiword expressions}},
  chapter~4. Language Science Press., Berlin.

\bibitem[{Savary and Cordeiro(2017)}]{Savary2017a}
Savary, Agata and Silvio~Ricardo Cordeiro. 2017.
\newblock Literal readings of multiword expressions: as scarce as hen{'}s
  teeth.
\newblock In \emph{Proceedings of the 16th International Workshop on Treebanks
  and Linguistic Theories}, pages 64--72, Prague, Czech Republic.

\bibitem[{Schneider et~al.(2016)Schneider, Hovy, Johannsen, and
  Carpuat}]{Schneider2016}
Schneider, Nathan, Dirk Hovy, Anders Johannsen, and Marine Carpuat. 2016.
\newblock {SemEval-2016 Task 10: Detecting Minimal Semantic Units and their
  Meanings (DiMSUM)}.
\newblock In \emph{Proceedings of {SemEval}}, pages 564--559.

\bibitem[{Senaldi, Lebani, and Lenci(2016)}]{Senaldi2016}
Senaldi, Marco S.~G., Gianluca~E. Lebani, and Alessandro Lenci. 2016.
\newblock Lexical variability and compositionality: Investigating idiomaticity
  with distributional semantic models.
\newblock In \emph{Proceedings of the 12th Workshop on {Multiword}
  {Expressions}}, pages 21--31.

\bibitem[{Seretan(2008)}]{Seret2008}
Seretan, Violeta. 2008.
\newblock \emph{Collocation Extraction Based on Syntactic Parsing}.
\newblock Ph.D. thesis, University of Geneva.

\bibitem[{Sporleder and Li(2009)}]{Sporleder2009}
Sporleder, Caroline and Linlin Li. 2009.
\newblock Unsupervised recognition of literal and non-literal use of idiomatic
  expressions.
\newblock In \emph{Proceedings of {EACL}}, pages 754--762.

\bibitem[{Sporleder et~al.(2010)Sporleder, Li, Gorinski, and
  Koch}]{Sporleder2010}
Sporleder, Caroline, Linlin Li, Philip~John Gorinski, and Xaver Koch. 2010.
\newblock Idioms in context: The {IDIX} corpus.
\newblock In \emph{Proceedings of {LREC}}, pages 639--646.

\bibitem[{Villavicencio(2005)}]{Villavicencio2005}
Villavicencio, Aline. 2005.
\newblock The availability of verb–particle constructions in lexical
  resources: How much is enough?
\newblock \emph{Computer Speech and Language}, 19(4):415--432.

\bibitem[{Williams et~al.(2015)Williams, Bannister, Arribas-Ayllon, Preece, and
  Spasi\'{c}}]{Williams2015}
Williams, Lowri, Christian Bannister, Michael Arribas-Ayllon, Alun Preece, and
  Irena Spasi\'{c}. 2015.
\newblock The role of idioms in sentiment analysis.
\newblock \emph{Expert Systems with Applications}, 42(21):7375--7385.

\end{thebibliography}

\end{document}